\documentclass[journal]{IEEEtran}

\usepackage{amsmath,amssymb,amsthm,mathtools,bm}
\usepackage{algorithm}
\usepackage{algorithmic}
\usepackage{graphicx}
\usepackage{booktabs}
\usepackage{multirow}
\usepackage{array}
\usepackage{tabularx}
\usepackage{placeins}
\usepackage{xspace}
\usepackage{xcolor}
\usepackage{tikz}
\usetikzlibrary{arrows.meta,positioning,fit,calc}
\usepackage{cite}
\usepackage{microtype}
\usepackage[hidelinks]{hyperref}

\AddToHook{cmd/normalsize/after}{%
  \setlength{\abovedisplayskip}{4pt plus 1pt minus 1pt}%
  \setlength{\belowdisplayskip}{4pt plus 1pt minus 1pt}%
  \setlength{\abovedisplayshortskip}{2pt plus 1pt minus 1pt}%
  \setlength{\belowdisplayshortskip}{3pt plus 1pt minus 1pt}%
}
\newtheorem{assumption}{Assumption}

\newtheorem{lemma}{Lemma}
\newtheorem{proposition}{Proposition}
\newtheorem{theorem}{Theorem}
\newtheorem{corollary}{Corollary}
\theoremstyle{remark}
\newtheorem{remark}{Remark}

\newcommand{\R}{\mathbb R}
\newcommand{\E}{\mathbb E}
\newcommand{\Pp}{\mathbb P}
\newcommand{\one}{\mathbf 1}

\newcommand{\col}{\operatorname{col}}
\newcommand{\supp}{\operatorname{supp}}
\newcommand{\Law}{\operatorname{Law}}
\newcommand{\Cov}{\operatorname{Cov}}

\newcommand{\abs}[1]{\left|#1\right|}
\newcommand{\argmaxop}{\operatorname*{arg\,max}}

\newcommand{\CPDST}{\textsc{CP-DST}\xspace}

\title{Decentralized Predictor--Corrector Sampling for\\
Offline Edge Optimization}

\author{Zai Shi, Bincheng Zhu, Kaikai Chi,~\IEEEmembership{Senior Member,~IEEE}, and Saiqin Long%
\thanks{Zai Shi, Bincheng Zhu, and Kaikai Chi are with the School of Computer Science and Technology, Zhejiang University of Technology, Hangzhou 310023, China (e-mail: shizai@zjut.edu.cn, bczhu@zjut.edu.cn, kkchi@zjut.edu.cn). \textit{(Corresponding author: Kaikai Chi.)}}%
\thanks{Saiqin Long is with the College of Information Science and Technology, Jinan University, Guangzhou 510632, China (e-mail: saiqinlong@jnu.edu.cn).}}

\begin{document}

\maketitle

\begin{abstract}
Decentralized offline optimization seeks high-utility configurations from heterogeneous historical data across edge nodes through local learning and neighbor exchanges, while keeping raw data and model parameters local. In this paper, we focus on communication cost at matched generation quality and common-support satisfaction. To this end, we propose calibrated product diffusion score tracking (\CPDST), a decentralized predictor--corrector sampler combining local noise-conditioned diffusion models trained by reward-weighted score matching. Its predictor calibrates transport between noise levels using local moment estimates, while score-tracked Langevin correction addresses residual mismatch through peer-to-peer exchanges.  For bounded design domains, we establish nonasymptotic Wasserstein bounds for non-log-concave targets, separating model, network, and discretization errors and linking sampling accuracy to communication budgets for offline-optimality guarantees under a coverage condition. Beyond fixed-target guarantees, we derive correction-cost lower bounds for heterogeneous bounded local factors. At prescribed levelwise accuracies, these bounds show when sharing calibration across candidate chains reduces total communication. Experiments across multiple applications demonstrate improved solution quality and common-support satisfaction over baselines under communication constraints. Further experiments quantify calibration gains across candidate counts and dimensions at equal communication cost.
\end{abstract}


\section{Introduction}
\label{sec:introduction}

We study \emph{decentralized offline optimization}, where nodes jointly optimize a shared configuration using locally stored historical records and neighbor communication, without a central optimizer or new objective evaluations. Each node's records reflect its own environment and utility, so a configuration that performs well locally may yield poor overall performance. The goal is to combine this distributed evidence to maximize total utility while keeping data and models local.

We consider offline settings in which multiple nodes measure different components of a shared objective and retain their own historical records. In beamforming~\cite{shi2011wmmse}, receiving sites retain local channel--rate records and jointly maximize the sum of local rates. In edge offloading~\cite{xu2026mec}, servers retain local delay--energy records, and we minimize their aggregate weighted cost. In wind-farm control~\cite{floris2026software}, controllers retain power records for their turbine groups and jointly maximize total farm power. Local objectives are coupled: an upstream turbine's yaw adjustment can reduce its own output while increasing downstream production. These settings motivate nodes to learn their respective utilities from local logs and coordinate a shared configuration through neighbor communication. Aggregating the local utilities balances their competing effects, while offline learning uses existing measurements without repeated trials on operating systems.

In decentralized offline optimization, heterogeneous records lead to different data coverage across nodes. A candidate well supported at one node may fall outside another’s observed region, requiring extrapolation that can overestimate its local utility~\cite{trabucco2021coms}. We therefore optimize over the common data support, where every utility term has local evidence. We focus on communication efficiency because generating a batch of candidate solutions requires nodes to exchange messages repeatedly as they refine candidates and reach consensus. The traffic depends on how many refinement steps each candidate needs. We aim to reduce this traffic while maintaining generation quality and common-support satisfaction.

For the above setup, we propose calibrated product diffusion score tracking (\CPDST), with the following contributions:
\begin{itemize}
 \item We reformulate decentralized offline optimization as a generation problem, combining diffusion models learned from local data to generate high-utility solutions within the nodes’ common data support without exchanging raw data. We further introduce a strategy called moment calibration to reduce communication during generation, whose cost can be shared across multiple candidate solutions.
 \item We establish finite-time sampling bounds of our algorithm and connect sampling accuracy to offline utility under a coverage condition. We quantify the tracking advantage and finite-sample calibration error, and derive upper and lower correction-count bounds that identify sufficient conditions for calibration to reduce communication.
 \item We conduct experiments across multiple application scenarios, problem dimensions, and network sizes, demonstrating improved solution quality and high common-support satisfaction under limited communication budgets for \CPDST. Further results validate the benefits of calibration, particularly when generating many candidates in high-dimensional problems.
\end{itemize}

\section{Related Work}
\label{sec:related}

\subsection{Offline optimization}

Offline black-box optimization (BBO) learns from fixed input--value records through surrogate-based and generative approaches. \emph{Surrogate-based methods} fit an objective predictor and optimize its predictions. Conservative objective models (COMs)~\cite{trabucco2021coms} penalize high predictions outside the observed data. Recent methods learn to rank candidate designs from offline records~\cite{tan2025ranking}. Lyu et al.~\cite{lyu2026ranking} further adapt ranking to the distribution mismatch between historical records and promising designs. Yang et al.~\cite{yang2026spade} use a conditional diffusion model as a forward predictor, combining moment and ranking calibration with regularization based on proximity to observed designs.

\emph{Generative methods} learn to produce promising designs. Denoising diffusion optimization models (DDOM)~\cite{krishnamoorthy2023diffusion} condition generation on objective values and emphasize high-value training records. Recent diffusion language models adapt pretrained generators to design--label records and align generation with improved labels~\cite{sun2026diffusionbbo}. Complementing these approaches, Yao et al.~\cite{yao2025diversity} use distribution matching to preserve diversity among high-quality candidates. 

All the above methods focus on centralized optimization. Distributed offline policy optimization methods such as FedOPO and FEDORA~\cite{shen2023fedopo,rengarajan2024fedora} use server-coordinated learning to combine clients' historical experience into a policy for a shared decision process. Contrarily, our formulation maximizes the sum of distinct local utilities over the common data support. 
\subsection{Decentralized optimization and sampling}

Decentralized gradient descent~\cite{yuan2016dgd} combines local gradient updates with neighbor averaging. Gradient tracking~\cite{nedic2017diging} propagates gradient changes to estimate the aggregate optimization direction. Communication-focused methods also combine compressed neighbor exchanges with gossip and error compensation for online optimization~\cite{yang2026compressed}. These methods obtain gradients or feedback from the objective. Our nodes learn update directions from historical records.

Decentralized Langevin methods~\cite{gurbuzbalaban2021desgld} combine neighbor mixing, local potential gradients, and random noise. Subsequent methods introduce consensus correction~\cite{islam2024extra}, share tracked gradient estimates to handle heterogeneous data~\cite{bhar2025tracking}, and incorporate gradient tracking into stochastic gradient Langevin dynamics (SGLD) over changing networks~\cite{bajwa2025diging}. A recent proximal extension~\cite{islam2026proximal} handles convex sampling constraints through shared regularization. These samplers use local potential gradients. We learn local scores from offline records to access the product target.

Decentralized diffusion models train experts independently and combine them through a router~\cite{mcallister2025ddm}, with recent extensions supporting heterogeneous training objectives~\cite{jiang2026hddm}. Our method combines local utility preferences into a common optimization target and coordinates generation through neighbor exchanges.

\section{System Model and Problem Formulation}
\label{sec:model}

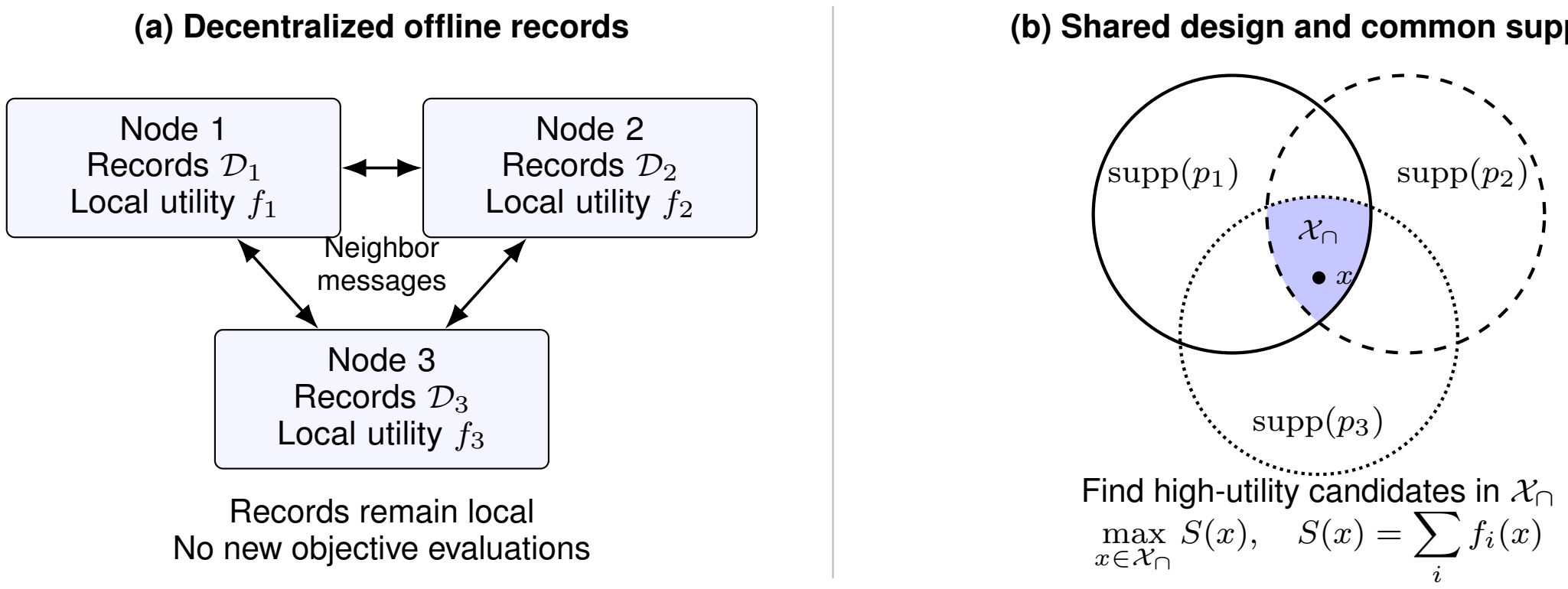
\begin{figure*}[t]
\centering
\begin{tikzpicture}[font=\sffamily\footnotesize,
 localnode/.style={draw,rounded corners=2pt,fill=blue!4,align=center,text width=2.65cm,minimum height=12mm},
 peer/.style={<->,>=Latex,line width=.7pt},
 title/.style={font=\sffamily\bfseries\footnotesize}]
\node[title] at (3.3,3.1) {(a) Decentralized offline records};
\node[localnode] (a) at (1.5,1.9) {Node 1\\Records $\mathcal D_1$\\Local utility $f_1$};
\node[localnode] (b) at (5.1,1.9) {Node 2\\Records $\mathcal D_2$\\Local utility $f_2$};
\node[localnode] (c) at (3.3,-.1) {Node 3\\Records $\mathcal D_3$\\Local utility $f_3$};
\draw[peer] (a) -- (b);
\draw[peer] (a) -- (c);
\draw[peer] (b) -- (c);
\node[align=center,font=\sffamily\scriptsize] at (3.3,1.05) {Neighbor\\messages};
\node[align=center] at (3.3,-1.25) {Records remain local\\No new objective evaluations};
\draw[gray!55] (7.2,-1.65) -- (7.2,3.3);
\node[title] at (11.4,3.1) {(b) Shared design and common support};
\begin{scope}
\clip (10.65,1.5) circle (1.2);
\clip (12.15,1.5) circle (1.2);
\fill[blue!22] (11.4,.45) circle (1.2);
\end{scope}
\draw[thick] (10.65,1.5) circle (1.2);
\draw[thick,dashed] (12.15,1.5) circle (1.2);
\draw[thick,densely dotted] (11.4,.45) circle (1.2);
\node at (10.15,1.85) {$\supp(p_1)$};
\node at (12.65,1.85) {$\supp(p_2)$};
\node at (11.4,-.3) {$\supp(p_3)$};
\node[font=\sffamily\scriptsize] at (11.4,1.35) {$\mathcal X_\cap$};
\fill (11.4,.95) circle (1.6pt);
\node[right,font=\sffamily\scriptsize] at (11.42,.95) {$x$};
\node[align=center] at (11.4,-1.25) {Find high-utility candidates in $\mathcal X_\cap$\\$\displaystyle\max_{x\in\mathcal X_\cap} S(x),\quad S(x)=\sum_i f_i(x)$};
\end{tikzpicture}
\caption{Decentralized offline optimization, illustrated with three nodes. Each node holds different design--utility records. Neighbor communication coordinates a shared design, whose total utility combines all local utilities and which must lie in the common support $\mathcal X_\cap=\bigcap_i\supp(p_i)$.}
\label{fig:problem_formulation}
\end{figure*}

\subsection{Edge nodes, private records, and common support}

We consider $n$ peer-to-peer edge nodes (e.g., access points, servers, or mobile devices) and designs $x\in\mathcal X\subseteq\R^d$.  Node $i$ has behavior density $p_i$, utility $f_i$, and $N_i$ noisy records:
\begin{equation*}
\begin{aligned}
 \mathcal D_i&=\{(x_{i,m},r_{i,m})\}_{m=1}^{N_i},
 \qquad x_{i,m}\sim p_i,\\
 r_{i,m}&=f_i(x_{i,m})+\nu_{i,m}.
\end{aligned}
\end{equation*}
Here $\nu_{i,m}$ denotes reward-label noise.  Both behavior distributions and utilities may differ across nodes.  With $\supp(p_i)$ denoting support, we define
$\mathcal X_\cap=\bigcap_{i=1}^{n}\supp(p_i)$ and
$S(x)=\sum_{i=1}^{n}f_i(x)$.
The common support provides local evidence for every utility term. The population objective is
\begin{equation*}
 x^\star\in\argmaxop_{x\in\mathcal X_\cap}S(x).
\end{equation*}
Nodes optimize through local models and neighbor messages without exchanging their raw data. We focus on the communication required to generate high-utility candidates within the common support. Centralized evaluation supplies the reported performance metrics. Figure~\ref{fig:problem_formulation} illustrates our problem setup.

\subsection{Communication model}
\label{subsec:communication_graph}

At communication round $r$, nodes exchange information over an undirected graph $\mathcal G_r=(\mathcal V,\mathcal E_r)$, where $\mathcal V=\{1,\ldots,n\}$. The mixing matrix $W_r\in\R^{n\times n}$ specifies how each node combines its own vector, e.g., $z_i$, with those received from its neighbors: $\sum_{j=1}^n [W_r]_{ij}z_j$.
Its entries are nonnegative, and $[W_r]_{ij}=0$ whenever $i\ne j$ and $\{i,j\}\notin\mathcal E_r$. We require $W_r$ to be doubly stochastic, so each update is a weighted average and the network average is preserved.

We denote the averaging matrix by $J=\mathbf 1\mathbf 1^\top/n$, where $\mathbf 1\in\R^n$ is the all-ones vector. For consecutive rounds, we write $W_{b:a}=W_b\cdots W_a$, with empty products equal to the identity matrix $I_n$. The spectral norm $\|W_{b:a}-J\|_2$ measures how far these mixing rounds remain from exact network averaging.

\begin{assumption}[Mixing and joint connectivity]
	\label{ass:graph}
	There exist an integer $B\ge1$ and a constant
	$\rho_B\in[0,1)$ such that
	\begin{equation*}
		\left\|W_{r+B-1:r}-J\right\|_2\le\rho_B,
	\end{equation*}
	for all $r\ge0$.
\end{assumption}

The window $B$ specifies a contraction timescale and allows disconnected individual graphs.  Nonnegative doubly stochastic matrices have spectral norm at most one and commute with $J$.  Partitioning any $m$-round block into $B$-round windows and a nonexpansive remainder gives
\begin{equation*}
 \|W_{r+m-1:r}-J\|_2\le\rho_B^{\lfloor m/B\rfloor},
 \qquad m\ge0,
\end{equation*}
with the right-hand side defined as one when $m<B$.

Denote $\mathcal N_i(r)$ as node $i$'s neighbor set and $d_i(r)=|\mathcal N_i(r)|$ as its degree. One construction uses Metropolis weights~\cite{nedic2017diging}:
\begin{equation*}
	[W_r]_{ij}=
	\begin{cases}
		\displaystyle
		\frac{1}{1+\max\{d_i(r),d_j(r)\}},
		& \{i,j\}\in\mathcal E_r,\\[2pt]
		\displaystyle
		1-\sum_{j'\in\mathcal N_i(r)}[W_r]_{ij'},
		& i=j,\\[2pt]
		0, & \text{otherwise},
	\end{cases}
\end{equation*}
These weights require only neighboring degree information
and are symmetric and doubly stochastic, with every
positive weight at least
$1/n$.
If the union graph
$(\mathcal V,\bigcup_{a=r}^{r+B-1}\mathcal E_a)$
is connected for every $r$, this construction ensures
the uniform contraction required by Assumption~\ref{ass:graph}. For a fixed graph, we write $W_r=W$ for all $r$.

\section{A generation view of the problem}
\label{sec:generation}

\subsection{Product-Gibbs target}

For inverse temperature $\beta>0$, node $i$ defines
\begin{equation*}
 q_{\beta,i}(x)=\frac{p_i(x)e^{\beta f_i(x)}}{\mathcal Z_i},
 \qquad \mathcal Z_i=\int p_i(u)e^{\beta f_i(u)}du.
\end{equation*}
We assume $0<\mathcal Z_i<\infty$ for every node (as ensured by bounded utilities on local supports), $0<\int\prod_i p_i(x)dx<\infty$, and $0<\int\prod_i q_{\beta,i}(x)dx<\infty$.  The global product is
\begin{align*}
 q_{\beta,G}(x)
 &=\frac{\prod_i q_{\beta,i}(x)}{\int\prod_i q_{\beta,i}(u)du} \\
 &\propto p_\cap(x)e^{\beta S(x)},
 \quad
 p_\cap(x)=\frac{\prod_i p_i(x)}{\int\prod_i p_i(u)du}.
\end{align*}
For a measurable set $A$, we write $p_\cap(A)=\int_A p_\cap(x)\,dx$.
Intuitively, $p_\cap$ restricts generation to designs supported by every node, while $e^{\beta S(x)}$ favors larger total utility. Increasing $\beta$ strengthens this preference, concentrating probability near the best designs within the common support. Sampling from this target thus turns the search for an optimum into the generation of high-utility candidates.

\subsection{Local diffusion models}
\label{subsec:diffusion_models}

Each local diffusion model learns a noise-conditioned score from labeled records~\cite{song2021sde}.  For $t\in[\varepsilon,T]$, $\varepsilon>0$, Gaussian corruption with density $K_t(u\mid x)$ uses signal scale $\alpha_t$, noise deviation $\sigma_t$, and independent standard Gaussian $\epsilon$:
\begin{equation*}
 U_t=\alpha_t X+\sigma_t\epsilon,\quad
 \epsilon\sim\mathcal N(0,I_d),\quad X\sim q_{\beta,i}.
\end{equation*}
The noised factor $q_{\beta,i,t}(u)=\int K_t(u\mid x)q_{\beta,i}(x)dx$ has score $s_i(u,t)=\nabla_u\log q_{\beta,i,t}(u)$.
The model $s_{\theta_i}:\R^d\times[\varepsilon,T]\to\R^d$, with local parameters $\theta_i$, gives
$\widehat s_i(u,t):=s_{\theta_i}(u,t)$.

For $r_i^{\max}=\max_m r_{i,m}$, reward weights are
\begin{equation*}
 a_{i,m}=\frac{\exp\{\beta(r_{i,m}-r_i^{\max})\}}
 {\sum_{m'=1}^{N_i}\exp\{\beta(r_{i,m'}-r_i^{\max})\}},
\end{equation*}
With noiseless labels, these weights implement self-normalized importance sampling from $p_i$ to $q_{\beta,i}$~\cite[Eq.~(3)]{paananen2021importance}, since $q_{\beta,i}/p_i\propto e^{\beta f_i}$, emphasizing high-utility designs while retaining the behavior distribution. Subtracting the maximum improves numerical stability. The conditional score is
\begin{equation*}
 \psi_t(u,x):=\nabla_u\log K_t(u\mid x)
 =-\frac{u-\alpha_tx}{\sigma_t^2}.
\end{equation*}
With fresh $t\sim\mathrm{Unif}[\varepsilon,T)$ and $\epsilon\sim\mathcal N(0,I_d)$, we set $u_{i,m,t}=\alpha_tx_{i,m}+\sigma_t\epsilon$.  For a noise-level weight $\lambda(t)>0$, node $i$ trains its model by minimizing
\begin{equation}
\begin{aligned}
 \widehat{\mathcal L}_i(\theta_i)
 =\sum_{m=1}^{N_i}a_{i,m}\E_{t,\epsilon}\big[
 &\lambda(t)\sigma_t^2\,
 \|s_{\theta_i}(u_{i,m,t},t)\\
 &\qquad-\psi_t(u_{i,m,t},x_{i,m})\|^2\big].
\end{aligned}
 \label{eq:weighted_score_matching}
\end{equation}
The factor $\sigma_t^2$ normalizes the conditional-score scale.

With noiseless labels, the unrestricted population minimizer is the local marginal score $s_i$, by the loss decomposition~\eqref{eq:score_loss_decomposition} in Appendix~\ref{app:preliminaries}.

The local noised distributions define a sequence of intermediate targets through their product:
\begin{equation*}
 \widetilde\pi_t(x)=\widetilde Z_t^{-1}
 \prod_{i=1}^{n}q_{\beta,i,t}(x),
\end{equation*}
where $\widetilde Z_t=\int\prod_i q_{\beta,i,t}(x)\,dx$. Its score $s_{\widetilde\pi_t}=\nabla\log\widetilde\pi_t$ decomposes over nodes:
$s_{\widetilde\pi_t}(x)=\sum_{i=1}^{n}s_i(x,t)$.
We construct these products by smoothing each local factor before multiplication. At the zero-noise endpoint, $\alpha_0=1$ and $\sigma_0=0$ give $\widetilde\pi_0=q_{\beta,G}$. If $\alpha_T=0$, every local marginal at $T$ is $\mathcal N(0,\sigma_T^2 I_d)$, giving the tractable starting distribution
$\widetilde\pi_T=\mathcal N(0,\sigma_T^2 I_d/n)$,
which all nodes can sample from a shared public seed.

With the target and intermediate distributions specified, the remaining task is to move samples from the starting Gaussian toward $q_{\beta,G}$ using only local scores and neighbor communication. We follow the schedule $T=t_0>t_1>\cdots>t_L=\varepsilon>0$, with terminal smoothing error included in the analysis. The next section develops a calibrated predictor--corrector sampler for these transitions.

\section{Calibrated Product Diffusion Score Tracking}
\label{sec:method}

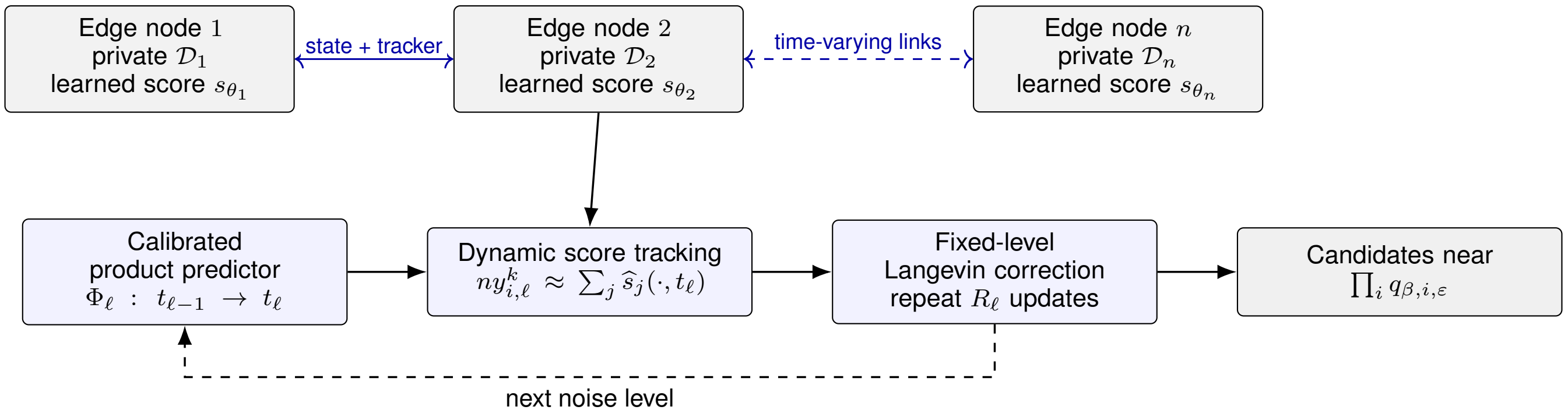
\begin{figure*}[t]
\centering
\begin{tikzpicture}[
 font=\sffamily\footnotesize,
 box/.style={draw,rounded corners=2pt,align=center,minimum height=10mm,inner sep=4pt},
 nodebox/.style={box,fill=gray!10,text width=3.0cm},
 stage/.style={box,text width=3.4cm},
 flow/.style={-{Latex[length=2.2mm]},line width=.65pt},
 peer/.style={<->,line width=.6pt,blue!65!black},
 peerlabel/.style={font=\sffamily\scriptsize,fill=white,inner sep=1.5pt,align=center}
]
\node[nodebox] (n1) {Edge node $1$\\private $\mathcal D_1$\\learned score $s_{\theta_1}$};
\node[nodebox,right=18mm of n1] (n2) {Edge node $2$\\private $\mathcal D_2$\\learned score $s_{\theta_2}$};
\node[nodebox,right=26mm of n2] (nn) {Edge node $n$\\private $\mathcal D_n$\\learned score $s_{\theta_n}$};
\draw[peer] (n1.east) -- node[peerlabel,above]{state + tracker} (n2.west);
\draw[peer,dashed] (n2.east) -- node[peerlabel,above]{time-varying links} (nn.west);

\node[stage,below=12mm of n1,xshift=4mm,fill=blue!5] (cal)
 {Calibrated product predictor\\$\Phi_\ell:\,t_{\ell-1}\to t_\ell$};
\node[stage,right=9mm of cal,fill=blue!5] (track)
 {Dynamic score tracking\\$ny_{i,\ell}^k\approx\sum_j\widehat s_j(\cdot,t_\ell)$};
\node[stage,right=9mm of track,fill=blue!5] (corr)
 {Fixed-level Langevin correction\\repeat $R_\ell$ updates};
\node[stage,right=9mm of corr,fill=gray!12] (out)
 {Candidates near\\$\prod_iq_{\beta,i,\varepsilon}$};
\draw[flow] (cal) -- (track);
\draw[flow] (track) -- (corr);
\draw[flow] (corr) -- (out);
\draw[flow] (n2.south) -- (track.north);
\draw[flow,dashed] (corr.south) -- ++(0,-6mm) -| node[below,pos=.25]{next noise level} (cal.south);
\end{tikzpicture}
\caption{\CPDST alternates prediction and correction across noise levels, exchanging states, score trackers, and optional calibration statistics. Data and trained models remain local.}
\label{fig:framework}
\end{figure*}

Figure~\ref{fig:framework} and Algorithm~\ref{alg:cpdst} summarize \CPDST, with the \textsc{Mix} subroutine defined in Algorithm~\ref{alg:mix}. After freezing models, node $i$ stores level-$\ell$ state $x_{i,\ell}^k$, tracker $y_{i,\ell}^k$, and cached score $c_{i,\ell}^k=\widehat s_i(x_{i,\ell}^k,t_\ell)$ after $k$ corrections. The affine predictor $\Phi_\ell$ initializes $k=0$ from level $\ell-1$, followed by $R_\ell$ corrections with step $\eta_\ell$. We enable calibration throughout by default (Remark~\ref{rem:shared_calibration}), with the flag fixed before inference.

\subsection{Diagonal product calibration}
\label{subsec:calibration}
As noise decreases, the target's center and coordinate variances change. Lines~\ref{alg:moment_cache} and~\ref{alg:calibrate} of Algorithm~\ref{alg:cpdst} estimate these changes to shift and rescale all $Q$ chains with one map per level. Diagonal moments require only two $d$-vectors per node. Score-based corrections then refine the remaining distributional shape.

Suppose that node $i$ has known design-radius bound $R_i>0$, with $B(0,R_i)=\{x:\|x\|\le R_i\}$.  From $M$ independent local samples, generated independently of inference and projected onto this ball, each node computes the empirical mean $\widehat u_i$, and coordinate-wise variances
$\widehat V_i$ with normalization $1/M$. Their corrupted moments are
\begin{equation*}
 \widehat\mu_{i,\ell}=\alpha_{t_\ell}\widehat u_i,\qquad
 \widehat D_{i,\ell}=\alpha_{t_\ell}^2\widehat V_i+\sigma_{t_\ell}^2I_d.
\end{equation*}
Since $\widehat D_{i,\ell}\succeq\sigma_{t_\ell}^2I_d\succ0$, we define the precision (inverse covariance) and information vector
\begin{equation*}
 \widehat\Lambda_{i,\ell}=\widehat D_{i,\ell}^{-1},
 \qquad
 \widehat h_{i,\ell}=\widehat\Lambda_{i,\ell}\widehat\mu_{i,\ell}.
\end{equation*}
Replacing local factors by $\mathcal N(\widehat\mu_{i,\ell},\widehat D_{i,\ell})$ gives the normalized product $g_\ell=\mathcal N(\widehat m_\ell,\widehat C_\ell)$, where
\begin{equation}
 \widehat C_\ell=\left(\sum_i\widehat\Lambda_{i,\ell}\right)^{-1},\qquad
 \widehat m_\ell=\widehat C_\ell\sum_i\widehat h_{i,\ell}.
 \label{eq:product_gaussian}
\end{equation}
Preserving the standardized coordinate $\widehat C_{\ell-1}^{-1/2}(x-\widehat m_{\ell-1})$ across levels gives
\begin{equation}
 \Phi_\ell(x)=\widehat m_\ell+\widehat A_\ell(x-\widehat m_{\ell-1}),\quad
 \widehat A_\ell=\widehat C_\ell^{1/2}\widehat C_{\ell-1}^{-1/2}.
 \label{eq:calibrated_map}
\end{equation}
Under exact aggregation, the map sends $g_{\ell-1}$ to $g_\ell$, and correction targets the residual mismatch. Finite consensus yields node-specific maps $\Phi_{i,\ell}$ as in~\eqref{eq:finite_calibration_consensus}, with sampling and consensus errors controlled by~\eqref{eq:calibration_sample_bound}--\eqref{eq:calibration_consensus_budget} in Appendix~\ref{app:prediction}.

Nodes aggregate the $d$ diagonal precision entries and $d$ information-vector entries by neighbor averaging with $W_r$, multiplying the mixed averages by $n$ to estimate the sums in~\eqref{eq:product_gaussian}. Algorithm~\ref{alg:cpdst} places this once-per-level aggregation outside the chain loop.

\subsection{Dynamic score tracking}
\label{subsec:why_tracking}

Algorithm~\ref{alg:cpdst}, lines~\ref{alg:corrector_cache}--\ref{alg:score_innovation}, adapts gradient tracking's mixing-and-innovation mechanism~\cite{nedic2017diging} to frozen scores. Here $m_\ell$ is the correction block length and $r_{\ell,k}$ its first round at level $\ell$, iteration $k$. We write $P_{\ell,k}=W_{r_{\ell,k}+m_\ell-1:r_{\ell,k}}$ and $\widehat b_{i,\ell}^k=ny_{i,\ell}^k$. With fresh $\xi_{\ell,k}\sim\mathcal N(0,I_d)$ shared by the nodes of each chain, the updates are
\begin{align*}
 x_{i,\ell}^{k+1}&=\sum_{j=1}^n[P_{\ell,k}]_{ij}x_{j,\ell}^k
 +\eta_\ell\widehat b_{i,\ell}^k
 +\sqrt{2\eta_\ell}\,\xi_{\ell,k},
 \\
 y_{i,\ell}^{k+1}&=\sum_{j=1}^n[P_{\ell,k}]_{ij}y_{j,\ell}^k
 +\widehat s_i(x_{i,\ell}^{k+1},t_\ell)
-\widehat s_i(x_{i,\ell}^k,t_\ell).
\end{align*}
Tracking score increments lets each node reuse the previous aggregate as candidates move. Double stochasticity and the score increment preserve
$\bar y_\ell^k=n^{-1}\sum_i\widehat s_i(x_{i,\ell}^k,t_\ell)$, where $\bar y_\ell^k=n^{-1}\sum_i y_{i,\ell}^k$ (Lemma~\ref{lem:average_invariant}).  Shared noise preserves the average's Langevin variance without adding disagreement.

On a fixed graph, $P_{\ell,k}=W^{m_\ell}$.  On time-varying graphs, the block-product bound in Section~\ref{subsec:communication_graph} gives
$\|P_{\ell,k}-J\|_2\le\gamma_\ell:=\rho_B^{\lfloor m_\ell/B\rfloor}<1$ when $m_\ell\ge B$. Theorem~\ref{thm:fixed_corrector} uses this contraction bound to analyze the correction loop.

\subsection{Level transition and tracker refresh}

The prediction phase in Algorithm~\ref{alg:cpdst}, lines~\ref{alg:transition_cache}--\ref{alg:predict_refresh}, mixes the previous level's states and trackers, applies the calibration map, and refreshes the tracker for the new noise level. Here $m_\ell^{\rm tr}$ is the transition block length and $r_\ell^{\rm tr}$ its first round. We set
$P_\ell^{\rm tr}=W_{r_\ell^{\rm tr}+m_\ell^{\rm tr}-1:r_\ell^{\rm tr}}$ and $\gamma_\ell^{\rm tr}=\|P_\ell^{\rm tr}-J\|_2$.
The resulting updates are
\begin{align*}
 x_{i,\ell}^0
 &=\Phi_{i,\ell}\!\left(\sum_j[P_\ell^{\rm tr}]_{ij}
 x_{j,\ell-1}^{R_{\ell-1}}\right),\\
 y_{i,\ell}^0
 &=\sum_j[P_\ell^{\rm tr}]_{ij}y_{j,\ell-1}^{R_{\ell-1}}
 +c_{i,\ell}^0-c_{i,\ell-1}^{R_{\ell-1}}.
\end{align*}
The refresh accounts for the change in both state and noise level, preserving the average-score identity. We choose $m_\ell^{\rm tr}\ge0$ separately from $m_\ell$, with $P_\ell^{\rm tr}=I_n$ for zero rounds and $\gamma_\ell^{\rm tr}\le\rho_B^{\lfloor m_\ell^{\rm tr}/B\rfloor}$ by the same block-product bound.

\begin{algorithm}[t]
\caption{\CPDST at node $i$ with $Q$ candidate chains}
\label{alg:cpdst}
\small
\renewcommand{\algorithmicrequire}{\textbf{Input:}}
\renewcommand{\algorithmicensure}{\textbf{Output:}}
\begin{algorithmic}[1]
\REQUIRE Local data/model, levels $t_{0:L}$, steps $\eta_{1:L}$, budgets $R_{1:L},m_{1:L},m_{1:L}^{\rm tr}$, weights $W_r$, calibration flag, $M$ calibration samples, and $Q$ candidate chains.
\ENSURE $Q$ candidates at $t_L=\varepsilon$.
\STATE Train and freeze $s_{\theta_i}$ by~\eqref{eq:weighted_score_matching}. Set $r=0$, $R_0=0$.
\STATE \label{alg:moment_cache} If calibrating, cache $(\widehat u_i,\widehat V_i)$ from $M$ independent projected samples, independent of the candidate chains. Aggregate once at $t_0$ for $(\widehat m_{i,0},\widehat C_{i,0})$.
\STATE Initialize each chain with $x_{i,0}^0\sim\widetilde\pi_T$ and $y_{i,0}^0=c_{i,0}^0=\widehat s_i(x_{i,0}^0,T)$.
\FOR{$\ell=1,\ldots,L$}
  \STATE \label{alg:calibrate} If calibrating, aggregate moment-derived statistics once and cache $\Phi_{i,\ell}$ and product moments via~\eqref{eq:product_gaussian}--\eqref{eq:calibrated_map}. Otherwise $\Phi_{i,\ell}(x)=x$.
  \FORALL{$Q$ chains in parallel (chain index omitted)}
    \STATE \textbf{Prediction phase}: reuse the same $\Phi_{i,\ell}$.
    \STATE \label{alg:transition_cache}\label{alg:transition_mix} $(u_{i,\ell}^{\rm tr},v_{i,\ell}^{\rm tr},r)\leftarrow
    \textsc{Mix}(x_{i,\ell-1}^{R_{\ell-1}},y_{i,\ell-1}^{R_{\ell-1}},m_\ell^{\rm tr},r)$.
    \STATE $x_{i,\ell}^0=\Phi_{i,\ell}(u_{i,\ell}^{\rm tr})$; $c_{i,\ell}^0=\widehat s_i(x_{i,\ell}^0,t_\ell)$.
    \STATE \label{alg:predict_refresh} $y_{i,\ell}^0=v_{i,\ell}^{\rm tr}+c_{i,\ell}^0-c_{i,\ell-1}^{R_{\ell-1}}$.
    \FOR{$k=0,\ldots,R_\ell-1$\label{alg:corrector_loop} (\textbf{Correction phase})}
      \STATE \label{alg:corrector_cache}\label{alg:mix_round} $(u_{i,\ell}^k,v_{i,\ell}^k,r)\leftarrow
      \textsc{Mix}(x_{i,\ell}^k,y_{i,\ell}^k,m_\ell,r)$.
      \STATE \label{alg:shared_noise} Draw shared $\xi_{\ell,k}\sim\mathcal N(0,I_d)$.
      \STATE \label{alg:state_update} $x_{i,\ell}^{k+1}=u_{i,\ell}^k+\eta_\ell n y_{i,\ell}^k+\sqrt{2\eta_\ell}\,\xi_{\ell,k}$.
      \STATE \label{alg:score_innovation} $c_{i,\ell}^{k+1}=\widehat s_i(x_{i,\ell}^{k+1},t_\ell)$; $y_{i,\ell}^{k+1}=v_{i,\ell}^k+c_{i,\ell}^{k+1}-c_{i,\ell}^k$.
    \ENDFOR
  \ENDFOR
\ENDFOR
\end{algorithmic}
\end{algorithm}

\begin{algorithm}[t]
\caption{\textsc{Mix}$(p_i,q_i,m,r)$ at node $i$}
\label{alg:mix}
\small
\renewcommand{\algorithmicrequire}{\textbf{Input:}}
\renewcommand{\algorithmicensure}{\textbf{Output:}}
\begin{algorithmic}[1]
\REQUIRE Local vectors $p_i,q_i\in\R^d$, integer budget $m\ge0$, global round $r$, available neighbors and weights.
\ENSURE Mixed vectors $u_i^m,v_i^m$ and advanced round $r$.
\STATE \label{alg:mix_init} Set $u_i^0=p_i$ and $v_i^0=q_i$.
\FOR{$a=1,\ldots,m$}
  \STATE \label{alg:mix_exchange} Exchange $(u_i^{a-1},v_i^{a-1})$ with $\mathcal N_i(r)$ and \label{alg:mix_average} simultaneously update from previous-round buffers:\\
  $\begin{aligned}
  u_i^a&=\textstyle\sum_{j\in\mathcal N_i(r)\cup\{i\}}[W_r]_{ij}u_j^{a-1},\\
  v_i^a&=\textstyle\sum_{j\in\mathcal N_i(r)\cup\{i\}}[W_r]_{ij}v_j^{a-1}.
  \end{aligned}$
  \STATE \label{alg:mix_advance} $r\leftarrow r+1$.
\ENDFOR
\STATE \textbf{return} $(u_i^m,v_i^m,r)$.
\end{algorithmic}
\end{algorithm}

Each chain has its own state and tracker. Initial draws and Gaussian increments are independent across chains and shared across nodes within each chain. With mixing matrices already chosen, each Gaussian increment is drawn independently of the full pre-update history. Parallel \textsc{Mix} calls share their input cursor and advance it once per batch. The incoming cursor gives $r_\ell^{\rm tr}$ or $r_{\ell,k}$ in the analysis.

Algorithm~\ref{alg:mix} uses call-local buffers and the same budget/cursor at all nodes. Calibration also advances $r$ each round.

\section{Theoretical Results}
\label{sec:analysis}

\subsection{Assumptions}

We write $\pi_t=\widetilde\pi_t$ and $b_t=\sum_i s_i(\cdot,t)$. With $\col$ denoting vector stacking, we define
$X_\ell^k=\col(x_{1,\ell}^k,\ldots,x_{n,\ell}^k)$,
$Y_\ell^k=\col(y_{1,\ell}^k,\ldots,y_{n,\ell}^k)$,
$\bar x_\ell^k=n^{-1}\sum_i x_{i,\ell}^k$, and $\Pi=I_n-J$.
Matrix actions on stacks use the Kronecker product with $I_d$.
We write $\widehat S_\ell(X_\ell^k)=\col(\widehat s_i(x_{i,\ell}^k,t_\ell))_{i=1}^n$ and $\|Z\|_{L^2}=(\E\|Z\|^2)^{1/2}$.  The unnormalized root-mean-square (RMS) disagreements are
$r_{x,\ell}^k=\|\Pi X_\ell^k\|_{L^2}$ and
$r_{y,\ell}^k=\|\Pi Y_\ell^k\|_{L^2}$.
We use $\Law$ for distributions and $W_2$ for quadratic Wasserstein distance with Euclidean ground distance. For $Z\sim\mu$, we write $\Phi_\#\mu=\Law(\Phi(Z))$. Our notation also uses $[z]_+=\max\{z,0\}$ and $\operatorname{Lip}(\Phi)$ for a Lipschitz constant.

\begin{assumption}[Bounded designs and positive noise]
\label{ass:support}
Each population factor has support in the design ball $B(0,R_i)$.  The finite schedule admits bounds $0\le\alpha_t\le\bar\alpha<\infty$ and $\sigma_t\ge\underline\sigma_\varepsilon>0$.
\end{assumption}

\begin{assumption}[Initial moments and score error]
\label{ass:scores}
The frozen score fields are Lipschitz. At each level $\ell=1,\ldots,L$, we assume finite initial moments and the following error bound at every analyzed correction step:
\begin{equation}
 \begin{gathered}
 \E\big[\|X_\ell^0\|^2+\|Y_\ell^0\|^2\big]<\infty,\\
 \|\widehat s_i(x_{i,\ell}^k,t_\ell)-s_i(x_{i,\ell}^k,t_\ell)\|_{L^2}
 \le\epsilon_{i,t_\ell},\quad
 \epsilon_{\Sigma,t_\ell}=\sum_i\epsilon_{i,t_\ell}.
 \end{gathered}
 \label{eq:score_error}
\end{equation}
\end{assumption}

Bound~\eqref{eq:score_error} accounts for data, approximation, training, and reward-label errors. A uniform-in-space score-error bound implies this condition.

\emph{Score architecture.}
We use $F_{\theta_i}$ for a multilayer perceptron with Lipschitz activations and finite layer spectral norms, and $e(t)$ a sinusoidal time embedding. We use a bounded residual to preserve confinement:
\begin{equation}
 \begin{aligned}
 s_{\theta_i}(u,t)&=-\frac{u}{\sigma_t^2}+g_{\theta_i}(u,t),
 \\
 g_{\theta_i}(u,t)&=\frac{\alpha_t R_i}{\sigma_t^2}
 \frac{F_{\theta_i}(u,e(t))}
 {\sqrt{1+\|F_{\theta_i}(u,e(t))\|^2}}.
\end{aligned}
\label{eq:bounded_score_residual}
\end{equation}
Thus $g_{\theta_i}(\cdot,t)$ is Lipschitz and $\|g_{\theta_i}(u,t)\|\le\alpha_tR_i/\sigma_t^2$.

To connect to the original optimization problem, we write
$S_\cap^\star=\max_{\mathcal X_\cap}S$.  For $\delta>0$, we define
$U_\delta=\{x:S(x)>S_\cap^\star-\delta\}$ and assume the coverage condition
\begin{equation}
 p_\cap(U_\delta)>0\quad\text{for every }\delta>0.
 \label{eq:optimum_coverage}
\end{equation}
This condition places positive product mass arbitrarily close to the common-support optimum.

\subsection{Convergence and offline optimality}
We first analyze correction at a fixed noise level $t_\ell$.

Using the cost $f_t$ in~\eqref{eq:cost_comparison} of Lemma~\ref{thm:harris_corrector}, we define
\begin{equation}
 E_{\ell,k}:=\inf_{\substack{\Law(U)=\Law(\bar x_\ell^k)\\
                            \Law(V)=\pi_{t_\ell}}}
              \E f_{t_\ell}(\|U-V\|).
 \label{eq:level_transport_cost}
\end{equation}
Here $k$ counts completed corrections. The transport costs $E_{\ell,0}$, $E_{\ell,R_\ell}$, and $E_{0,0}$ measure distribution error after prediction, after correction, and at initialization, respectively.

\begin{theorem}[Correction accuracy and tracking advantage]
	\label{thm:fixed_corrector}
	We condition on frozen models and calibration statistics independent
	of sampling randomness. Under Assumptions~\ref{ass:graph}--\ref{ass:scores}
	and architecture~\eqref{eq:bounded_score_residual},
	with $m_\ell\ge B$ and sufficiently small step size
	$\eta_\ell>0$, the level-$\ell$ corrections of
	Algorithm~\ref{alg:cpdst} satisfy:
	\begin{enumerate}
		\renewcommand{\labelenumi}{(\arabic{enumi})}
		\item With $E_{\ell,0}$ defined by~\eqref{eq:level_transport_cost}, every node satisfies
		\begin{equation}
			W_2(\Law(x_{i,\ell}^{R_\ell}),\pi_{t_\ell})
			\le r_{x,\ell}^{R_\ell}
			+\sqrt{\frac{
					q_\ell^{R_\ell}E_{\ell,0}+\Xi_\ell(R_\ell)}
				{\kappa_{-,t_\ell}}}.
			\label{eq:fixed_corrector_bound}
		\end{equation}
		The bound combines node disagreement, a geometrically decaying
		initial error, and accumulated discretization and score errors.
		Here $q_\ell=1-c_{t_\ell}\eta_\ell\in(0,1)$ is the contraction
		factor from~\eqref{eq:euler_cost_contraction},
		$\kappa_{-,t_\ell}>0$ is the comparison constant in~\eqref{eq:cost_comparison}
		(Appendix~\ref{app:preliminaries}), and $\Xi_\ell$ is defined
		by~\eqref{eq:corrector_cost_floor} in Appendix~\ref{app:proofs}.
		
		\item With diffusion noise level, models, mixing bound, and initial second-moment
		bounds fixed, tracking yields
		$\limsup_{k\to\infty}r_{x,\ell}^k=O(\eta_\ell^{3/2})$.
		Replacing the tracker $ny_{i,\ell}^k$ by the local score
		$n\widehat s_i(x_{i,\ell}^k,t_\ell)$ instead yields
		$\Omega(\eta_\ell)$ long-run RMS disagreement when the local
		scores satisfy~\eqref{eq:persistent_heterogeneity} with a
		step-size-independent lower bound. Both orders are taken as
		$\eta_\ell\to0$.
	\end{enumerate}
\end{theorem}

\emph{Proof.} See Appendices~\ref{app:proof_part1} and~\ref{app:proof_part2} for parts (1) and (2), respectively.

\begin{remark}[Vanishing correction error]
\label{rem:vanishing_correction}
For fixed $\beta$ and a finite positive-noise schedule, we retain Theorem~\ref{thm:fixed_corrector}'s hypotheses. With $\gamma_\ell\le\bar\gamma<1$ and uniformly bounded model Lipschitz constants and post-prediction state/tracker second moments, every node's $W_2$ error vanishes when, at each level,
$\eta_\ell\to0$, $\epsilon_{\Sigma,t_\ell}\to0$, and
$\eta_\ell R_\ell\to\infty$.
Tracking thus allows the communication rounds per correction to remain fixed.
\end{remark}
Theorem~\ref{thm:fixed_corrector} bounds correction error at one noise level. We next combine this bound with the predictor's effect to track error across the full schedule. We condition on the shared affine predictor $\Phi_\ell$ and define
\[
K_\ell=\operatorname{Lip}(\Phi_\ell),\qquad
b_\ell=W_2((\Phi_\ell)_\#\pi_{t_{\ell-1}},\pi_{t_\ell}).
\]
Here $K_\ell$ measures error amplification by the predictor,
and $b_\ell$ measures its mismatch between consecutive targets.
\begin{theorem}[Error control across noise levels]
\label{thm:end_to_end}
Under Theorem~\ref{thm:fixed_corrector}'s conditions at each level, we use a common affine predictor and define
\begin{align*}
 \mathcal A_\ell&=\frac{2\kappa_{+,t_\ell}}{\kappa_{-,t_{\ell-1}}}
             \max(K_\ell,K_\ell^2),\\
 \mathcal B_\ell&=2\kappa_{+,t_\ell}(b_\ell+b_\ell^2).
\end{align*}
With $E_{\ell,R_\ell}$ defined by~\eqref{eq:level_transport_cost},
\begin{equation*}
 E_{\ell,R_\ell}\le q_\ell^{R_\ell}(\mathcal A_\ell E_{\ell-1,R_{\ell-1}}+\mathcal B_\ell)
               +\Xi_\ell(R_\ell),
\end{equation*}
and, with empty products equal to one,
\begin{equation}
 \begin{aligned}
 E_{L,R_L}\le{}&\left(\prod_{j=1}^L q_j^{R_j}\mathcal A_j\right)E_{0,0}\\
 &+\sum_{\ell=1}^L\left(\prod_{j=\ell+1}^L q_j^{R_j}\mathcal A_j\right)
       [q_\ell^{R_\ell} \mathcal B_\ell+\Xi_\ell(R_\ell)].
 \end{aligned}
 \label{eq:end_to_end}
\end{equation}
For each node $i$, the final distribution satisfies
$W_2(\Law(x_{i,L}^{R_L}),\pi_\varepsilon)\le r_{x,L}^{R_L}+\sqrt{E_{L,R_L}/\kappa_{-,t_L}}$.
Choosing $R_\ell$ so that $q_\ell^{R_\ell}\mathcal A_\ell<1$ contracts the inherited-error term in the bound. $\mathcal B_\ell$ bounds the added transport cost from prediction mismatch.  
\end{theorem}

\emph{Proof.} See Appendix~\ref{app:finite_consensus}.

\begin{remark}[Propagation across levels]
\label{rem:vanishing_levels}
For fixed $L$, Remark~\ref{rem:vanishing_correction}'s conditions and uniformly bounded $\mathcal A_\ell,\mathcal B_\ell$ imply $E_{L,R_L}\to0$, using the adjusted coefficients~\eqref{eq:finite_consensus_coefficients} in Appendix~\ref{app:finite_consensus} for node-specific maps. Sufficient correction attenuates the contribution of bounded prediction and calibration errors, so the numbers of communication rounds for mixing states and trackers between noise levels and aggregating calibration statistics may remain fixed.
\end{remark}

We next convert Theorem~\ref{thm:end_to_end}'s sampling error and terminal smoothing error into a utility guarantee.

\begin{theorem}[Offline value guarantee]
\label{thm:optimality}
For node $i$'s output distribution $\mu_i$, we assume~\eqref{eq:optimum_coverage} holds and $S$ is $L_S$-Lipschitz on the output domain, or is given a specified Lipschitz extension.  If $W_2(\mu_i,q_{\beta,G})\le\Delta_i$, then for every $\delta>0$,
\begin{equation}
 \Pp_{X\sim\mu_i}\{S(X)\le S_\cap^\star-\delta\}
 \le\frac{e^{-\beta\delta/4}}{p_\cap(U_{\delta/4})}
      +\frac{4L_S^2\Delta_i^2}{\delta^2}.
 \label{eq:sampling_to_value}
\end{equation}
With terminal smoothing error $\delta_{\rm sm}(\varepsilon)=W_2(\pi_\varepsilon,q_{\beta,G})$, we may choose the error bound
\begin{equation*}
 \Delta_i=r_{x,L}^{R_L}+\sqrt{E_{L,R_L}/\kappa_{-,t_L}}
               +\delta_{\rm sm}(\varepsilon).
\end{equation*}
For bounded compactly supported local densities with positive product integral, $\alpha_\varepsilon\to1$ and $\sigma_\varepsilon\to0$ imply $\delta_{\rm sm}(\varepsilon)\to0$ at fixed $\beta$.
If $\mathcal X_\cap$ is known and admits an implementable measurable nearest-point projection, the projected output is feasible and satisfies~\eqref{eq:sampling_to_value} with $2\Delta_i$ in place of $\Delta_i$, or $\Delta_i$ if $\mathcal X_\cap$ is convex.
\end{theorem}

\emph{Proof.} See Appendix~\ref{app:offline_value}.

\begin{remark}[Vanishing value gap]
\label{rem:vanishing_value}
Under Theorem~\ref{thm:optimality}'s hypotheses for a fixed optimization problem and $\delta>0$, the failure probability vanishes when $\beta\to\infty$, $\Delta_i\to0$, and any calibration failure probability tends to zero. We choose terminal noise and sampling tolerances for each finite $\beta$ using the fixed-$\beta$ smoothing limit in Theorem~\ref{thm:optimality}.
\end{remark}

For target gap $\delta>0$, prescribed failure probability $\nu\in(0,1)$, and $L_S>0$, sufficient conditions for~\eqref{eq:sampling_to_value} to be at most $\nu$ are
\begin{equation}
 \beta\ge\frac4\delta\log\frac{2}{\nu p_\cap(U_{\delta/4})},\qquad
 \Delta_i\le\frac{\delta\sqrt\nu}{2\sqrt2 L_S}.
 \label{eq:optimality_wasserstein_budget}
\end{equation}
\subsection{Communication cost}
\label{subsec:predictor_value}

In this part, we analyze the communication cost of our algorithm. We count scalar entries transmitted in both directions over neighbor links. On a fixed graph with edge set $\mathcal E$, each mixing round sends two $d$-vectors per chain, while calibration sends $2d$ statistics shared by all chains. With $m_\ell^{\rm cal}$ calibration rounds at level $\ell$, the state/tracker and calibration costs are
\begin{align}
 N_{\rm scal}^{x,y}
 &=4Qd\abs{\mathcal E}\sum_{\ell=1}^{L}(m_\ell R_\ell+m_\ell^{\rm tr}),
 \label{eq:multichain_state_traffic}\\
 N_{\rm scal}^{\rm cal}
 &=4d\abs{\mathcal E}\sum_{\ell=0}^{L}m_\ell^{\rm cal}.
 \label{eq:shared_calibration_traffic}
\end{align}
The total cost is the sum of~\eqref{eq:multichain_state_traffic} and~\eqref{eq:shared_calibration_traffic}. We write $c_\ell^{\rm step}=4d|\mathcal E|m_\ell$ for one correction's cost per chain and $C^{\rm cal}=N_{\rm scal}^{\rm cal}$ for the shared calibration cost. Cached initial statistics give $m_0^{\rm cal}=0$. For time-varying graphs, we count the realized edges in each round.

We compare calibration with identity prediction ($\Phi_{i,\ell}(x)=x$, $m_\ell^{\rm cal}=0$), retaining mixing and tracker refresh. Both policies share the fixed graph, step sizes, mixing budgets, local factors, reward temperature, schedule, and each chain's initial state. At every level $\ell$, each node and the network average must have $W_2$ error at most $\tau_\ell>0$ relative to $\pi_{t_\ell}$. The initial network-average error is at most $\tau_0\ge0$ relative to $\pi_{t_0}$. For $Q$ candidate chains, $C_{\rm on}(Q)$ denotes the minimum total scalar traffic among calibrated runs meeting these tolerances, varying only the correction counts under the fixed settings above. We write $C_{\rm off}(Q)$ for the total scalar traffic of any uncalibrated run meeting the same tolerances.

\begin{theorem}[When calibration saves communication]
\label{thm:predictor_separation}
The following bound holds with probability at least $1-\zeta$ over the calibration samples, provided that both policies satisfy Theorem~\ref{thm:fixed_corrector} and $0<n\eta_\ell/\sigma_{t_\ell}^2<1$ at each level:
\begin{equation}
 \begin{aligned}
 &C_{\rm off}(Q)-C_{\rm on}(Q)\ge Q\Delta_{\rm corr}-C^{\rm cal},\\
 &\Delta_{\rm corr}:=\sum_{\ell=1}^L c_\ell^{\rm step}
 (\underline R_\ell^{\rm off}-\overline R_\ell^{\rm on}).
 \end{aligned}
 \label{eq:instance_break_even}
\end{equation}
Here $\overline R_\ell^{\rm on}<\infty$ and $\underline R_\ell^{\rm off}$ are defined in~\eqref{eq:predictor_tolerance_test} and~\eqref{eq:full_run_lower_counts}, respectively, with
\begin{equation*}
\begin{aligned}
 v_\ell&=\widehat K_\ell\tau_{\ell-1}+B_\ell+e_\ell^{\rm cal}+e_\ell^{\rm net},\\
 \Delta_{\ell,0}&=[\|\mu_{t_{\ell-1}}-\mu_{t_\ell}\|-\tau_{\ell-1}]_+,\\
 \omega_\ell&=\alpha_{t_\ell}r_{\rm av}+\|\mu_{t_\ell}-\alpha_{t_\ell}c\|
 +\epsilon_{\Sigma,t_\ell}/a_\ell.
\end{aligned}
\end{equation*}
Here $\widehat K_\ell=\operatorname{Lip}(\widehat\Phi_\ell)$ for the common estimated predictor, $B_\ell$ bounds ideal prediction mismatch, and $e_\ell^{\rm cal},e_\ell^{\rm net}$ measure calibration and finite-consensus errors. Also, $\mu_t=\E_{\pi_t}X$, $a_\ell=n/\sigma_{t_\ell}^2$, and $r_{\rm av}=n^{-1}\sum_i r_i$, where $\supp(q_{\beta,i})\subseteq B(c,r_i)$ for a common center $c$. We use $\mathbb N_0=\{0,1,\ldots\}$ and $\min\varnothing=+\infty$ in~\eqref{eq:predictor_tolerance_test}. If $\Delta_{\rm corr}>0$ and $Q>C^{\rm cal}/\Delta_{\rm corr}$, a calibrated run meets the same levelwise tolerances at lower cost than any uncalibrated run.
\end{theorem}

\emph{Proof.} See Appendix~\ref{app:prediction}.

\begin{remark}[Interpretation of the communication bound]
\label{rem:shared_calibration}
The costs $c_\ell^{\rm step}$ and $C^{\rm cal}$ depend on $d$, $|\mathcal E|$, and mixing/calibration rounds. At fixed tolerances, $\overline R_\ell^{\rm on}$ depends on prediction/calibration errors, node disagreement, mixing, step size, and score error. The lower bound $\underline R_\ell^{\rm off}$ depends on target mean shifts, support spread, step size, diffusion noise, and score error. Large target shifts increase correction needs, while accurate calibration reduces residual error. The configuration determines $Q$ and traffic costs. Applying these bounds requires estimates of target and score errors, and may yield conservative correction budgets. We therefore examine how calibration gains vary with candidate count and dimension through experiments in Section~\ref{subsec:calibration_validation}.
\end{remark}

\section{Experiments}
\label{sec:experiments}

We evaluate quality and coverage under communication limits on three applications, then study calibration on Rastrigin functions of different dimensions. We report the mean and standard error (SE) over five independent seeds, each controlling data, training, and a single sampling run. Appendix~\ref{app:protocol} provides additional details.

\subsection{Training, networks, and evaluation}
\label{subsec:score_training}

We train each local score model using~\eqref{eq:weighted_score_matching} and architecture~\eqref{eq:bounded_score_residual}, with two width-128 $\tanh$ layers and eight sinusoidal time features. Adam uses learning rate $10^{-3}$, batch size 128 per node, and 6,000 updates. Training times are uniform in $[0.02,0.98]$. The corruption schedule is $\alpha_t=\cos(\pi t/2)$ and $\sigma_t=\sin(\pi t/2)$.  All methods share local records, and score-based methods share frozen local models.

We use a ring to test communication limits: opposite nodes are $n/2$ hops apart. Throughout the run, nodes alternate neighbors each round, weighting their own and received vectors by 0.75 and 0.25. For $B=2$, the mixing matrices give $\rho_B=0.5$ for $n=4$ and $\rho_B\approx0.83113$ for $n=8$ (Assumption~\ref{ass:graph}). We count float32 payloads in both directions in mebibytes (MiB, $2^{20}$ bytes). The known initial Gaussian requires no calibration exchange. We evaluate unprojected outputs.

\begin{table}[t]
\centering
\caption{Node count $n$, input dimension $d$, and base communication budget $B_0$ (MiB/run).}
\label{tab:experiment_settings}
\begin{tabular}{lrrr}
\toprule
Task & $n$ & $d$ & $B_0$\\
\midrule
Beamforming & 8 & 8 & 1.125\\
Offloading & 4 & 24 & 1.688\\
FLORIS & 8 & 64 & 9.000\\
\bottomrule
\end{tabular}
\end{table}

\subsection{Application tasks and comparisons}
\label{subsec:edge_task}
\label{subsec:floris_task}

\emph{Beamforming.} Each of eight nodes is an edge controller at a receiving site. They jointly optimize a shared transmitter's eight-antenna beamforming vector, with $w_m(x)=e^{\mathrm j\pi x_m}/\sqrt8$ for antenna $m=1,\ldots,8$, where $\mathrm j^2=-1$. We maximize $S(x)=\sum_{i=0}^{7}f_i(x)$ over the common input support $\mathcal X_\cap$, where
\begin{equation}
\label{eq:experiment_beamforming_objective}
f_i(x)=\frac1{64}\sum_{s=1}^{64}\log\!\left(1+\frac{|h_{i,s}^Hw(x)|^2}{0.1}\right).
\end{equation}
Here $h_{i,s}\in\mathbb C^8$ is site $i$'s channel vector in snapshot $s$, $H$ denotes conjugate transpose, and 0.1 is the noise power. We report test $S(x)$ in nat/s/Hz.

Site $i$'s offline dataset contains 2,048 simulated pairs $(x,f_i(x))$. Each $x\in\R^8$ specifies antenna phases $\pi x_m$. Coordinates are independently drawn from $\mathcal N(0.15+0.15\sin(i),0.48^2)$, rejecting vectors outside $[-1+0.4i/7,0.6+0.4i/7]^2\times[-1,1]^6$. The label $f_i(x)$ is the average local rate over 64 fixed training channel snapshots, computed by~\eqref{eq:experiment_beamforming_objective}. The common input support is $[-0.6,0.6]^2\times[-1,1]^6$.

\emph{Offloading.} Node $i\in\{0,1,2,3\}$ is an edge server serving users $2i,2i+1$. The first, middle, and last eight coordinates of $x\in\R^{24}$ encode all users' offloading fractions, bandwidth shares, and CPU shares, respectively. We maximize $S(x)=\sum_{i=0}^{3}f_i(x)$ over $\mathcal X_\cap$, equivalently minimizing total delay plus weighted energy, with
\begin{equation}
\label{eq:experiment_offloading_objective}
f_i(x)=-\frac1{64}\sum_{s=1}^{64}\sum_{u=2i}^{2i+1}[T_{u,s}(x)+0.2E_{u,s}(x)].
\end{equation}
Here $T_{u,s}$ and $E_{u,s}$ are user $u$'s completion time and energy in workload $s$, defined by~\eqref{eq:experiment_offloading_cost}. We report test $S(x)$, so larger values mean lower cost.

Server $i$'s offline dataset contains 2,048 simulated pairs $(x,f_i(x))$. We draw independent $\mathcal N(0.15+0.15\sin(i),0.48^2)$ coordinates, rejecting vectors outside $[-1+0.4i/3,0.6+0.4i/3]^2\times[-1,1]^{22}$. Equation~\eqref{eq:experiment_offloading_mapping} in Appendix~\ref{app:protocol} maps $x$ to offloading fractions and resource shares. Labels use~\eqref{eq:experiment_offloading_objective} on 64 fixed lognormal training workloads. The common input support is $[-0.6,0.6]^2\times[-1,1]^{22}$.

\emph{Wind-farm yaw control (FLORIS).} We simulate 64 wind turbines using FLOw Redirection and Induction in Steady State (FLORIS)~\cite{floris2026software}. Eight local controllers each manage eight turbines in one column of an $8\times8$ array. The shared input $x\in\R^{64}$ specifies all turbine yaw angles as $30x_j$ degrees. Each controller values its own turbines' power, while the shared goal is to maximize total farm power over $\mathcal X_\cap$:
\begin{equation}
\label{eq:experiment_floris_objective}
S(x)=\sum_{i=0}^{7}f_i(x),\quad f_i(x)=\sum_{j=8i+1}^{8i+8}P_j(x).
\end{equation}
Here $P_j(x)$ is turbine $j$'s simulated power in MW under fixed wind conditions. Turning an upstream turbine can reduce its own power but redirect its wake, increasing downstream production. Thus controllers have distinct, coupled local utilities.

Each controller stores 2,048 simulated pairs $(x,f_i(x))$. Coordinates are independently drawn from $\mathcal N(0.15+0.15\sin(i),0.48^2)$, rejecting vectors outside $[-1+0.4i/7,0.6+0.4i/7]^2\times[-1,1]^{62}$. Labels use~\eqref{eq:experiment_floris_objective}. Each controller retains only its local power labels. The common input support is $[-0.6,0.6]^2\times[-1,1]^{62}$. After generating these offline records, the sampler uses the local records and fitted models. FLORIS evaluates the validation and test candidates. Appendix~\ref{app:protocol} gives the wind and turbine settings and the choice of $\beta$ via~\eqref{eq:experiment_floris_ess}.

Existing offline design methods typically assume pooled data. We compare \CPDST with four baselines that retain data locally to quantify the effects of calibration and score tracking. All use the same learned scores and identity prediction between levels. Tracked correction (Tracking) retains the correction step of \CPDST, mixing states and updating persistent score trackers through local-score changes. State mixing (State) exchanges only states and uses local scores for correction. Restarted score consensus (Restart) mixes states and then aggregates local scores from scratch at each correction. Local-only generation (Local) runs each node's sampler independently with zero communication.

\subsection{Communication caps and application results}
\label{subsec:tmc_protocol}

We measure each method's quality and coverage under communication budgets $\{0.5,1,2,4,6\}B_0$, with task-specific $B_0$ given in Table~\ref{tab:experiment_settings}. For each method and seed, we select a schedule on validation data from all configurations within the cap, including lower-budget schedules, then generate $Q=64$ candidates per node in one test run. For node $i$'s outputs $x_i^{(q)}$, the metrics are
\begin{equation}
\begin{aligned}
\mathrm{Quality}&=\min_i\max_{q:\,x_i^{(q)}\in\mathcal C}S(x_i^{(q)}),\\
\mathrm{Coverage}&=\min_i\frac1Q\sum_{q=1}^Q\mathbf1\{x_i^{(q)}\in\mathcal C\}.\label{eq:metrics}
\end{aligned}
\end{equation}
Here $\mathcal C=\mathcal X_\cap$ is each task's common input support, and $S$ is its total utility. Both metrics describe the weakest node. If any node has no covered candidate, the run fails.

Calibration fixes $M=1{,}024$ weighted local draws to estimate moments and 16 mixing rounds per level to reduce disagreement between nodes. Appendix~\ref{app:protocol} details the schedule search and validation criteria.

\begin{figure}[t]
\centering
\includegraphics[width=\columnwidth]{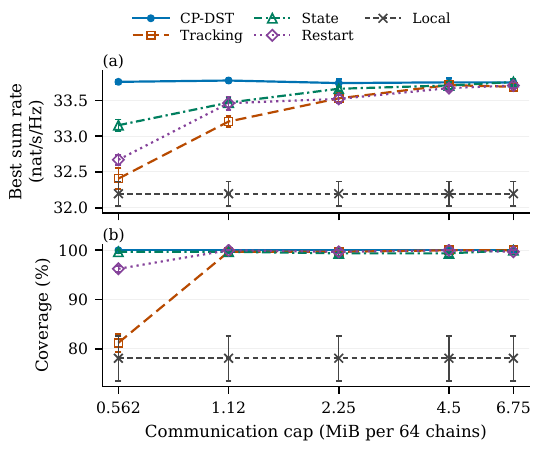}
\caption{Beamforming under communication caps: (a) worst-node quality and (b) minimum node coverage. Distinct markers and line styles identify methods. Error bars show SE over five seeds.}
\label{fig:beamforming}
\end{figure}
\begin{figure}[t]
\centering
\includegraphics[width=\columnwidth]{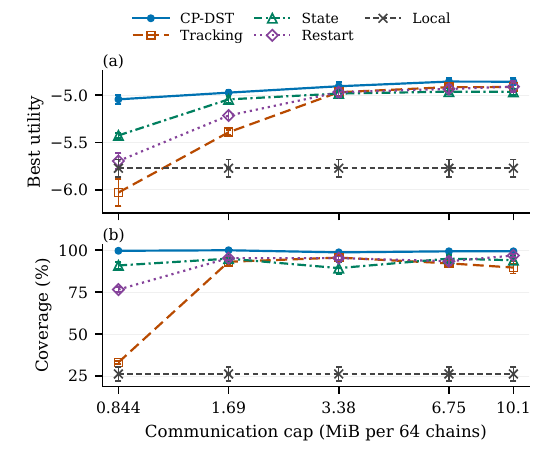}
\caption{Offloading: (a) worst-node utility and (b) minimum node coverage. Higher utility means lower weighted delay and energy. Error bars show SE over five seeds.}
\label{fig:offloading}
\end{figure}
\begin{figure}[t]
\centering
\includegraphics[width=\columnwidth]{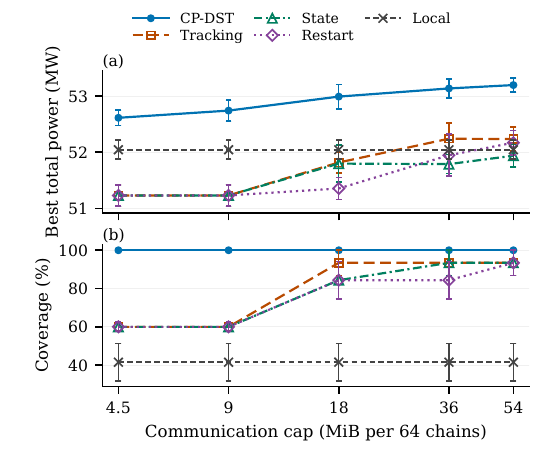}
\caption{FLORIS yaw control ($d=64$, $n=8$): (a) worst-node total power and (b) minimum node coverage, with SE over five seeds.}
\label{fig:floris}
\end{figure}

Figures~\ref{fig:beamforming}--\ref{fig:floris} show that \CPDST achieves the highest mean quality in 14 of the 15 task--budget combinations, while maintaining high coverage. The low-budget gains in quality and coverage demonstrate its effectiveness when communication is limited. At $0.5B_0$, \CPDST attains $33.763\pm0.028$ in beamforming and $-5.044\pm0.046$ in offloading, compared with $33.155\pm0.081$ and $-5.426\pm0.026$ for State, the strongest alternatives in quality. Its minimum node coverage is $100.00\pm0.00\%$ and $99.69\pm0.31\%$, respectively. 

For FLORIS (Fig.~\ref{fig:floris}), \CPDST achieves the highest mean quality at every budget, exceeding the strongest baseline by 1.09--1.84\%, with 100\% coverage throughout. At 4.5 MiB, its quality is $52.614\pm0.137$ MW, versus $52.045\pm0.171$ MW for Local. At 54 MiB, it reaches $53.196\pm0.127$ MW, versus $52.234\pm0.216$ MW and $93.44\pm6.56\%$ coverage for Tracking. Validation--test differences can produce nonmonotone test curves.

\begin{table*}[t]
\centering
\footnotesize
\setlength{\tabcolsep}{3pt}
\caption{Node variances averaged over five caps within each seed, then reported as mean $\pm$ SE over five seeds. Bold: lowest mean. Underline: second-lowest. Coverage uses fractions in $[0,1]$.}
\label{tab:node_agreement}
\begin{tabular}{llrrrrr}
\toprule
Task & Metric & CP-DST & Tracking & State & Restart & Local\\
\midrule
Beamforming & Quality & {\boldmath$(5.1\pm1.2)\!\times\!10^{-5}$} & $\underline{(1.3\pm0.2)\!\times\!10^{-3}}$ & $(2.8\pm0.2)\!\times\!10^{-3}$ & $(3.3\pm0.3)\!\times\!10^{-3}$ & $(1.4\pm0.4)\!\times\!10^{-1}$\\
 & Coverage & {\boldmath$0\pm0$} & $(1.9\pm0.8)\!\times\!10^{-5}$ & $(1.1\pm0.7)\!\times\!10^{-5}$ & $\underline{(8.1\pm4.3)\!\times\!10^{-6}}$ & $(5.8\pm2.4)\!\times\!10^{-3}$\\
\midrule
Offloading & Quality & {\boldmath$(2.7\pm1.2)\!\times\!10^{-5}$} & $(3.3\pm0.7)\!\times\!10^{-4}$ & $(7.2\pm1.1)\!\times\!10^{-4}$ & $\underline{(2.6\pm0.6)\!\times\!10^{-4}}$ & $(3.1\pm0.5)\!\times\!10^{-2}$\\
 & Coverage & {\boldmath$(2.4\pm2.4)\!\times\!10^{-6}$} & $(1.3\pm0.2)\!\times\!10^{-4}$ & $(2.2\pm1.3)\!\times\!10^{-4}$ & $\underline{(1.3\pm0.3)\!\times\!10^{-4}}$ & $(2.1\pm0.5)\!\times\!10^{-2}$\\
\midrule
FLORIS & Quality & {\boldmath$(10.0\pm0.9)\!\times\!10^{-3}$} & $\underline{(2.8\pm0.8)\!\times\!10^{-2}}$ & $(4.1\pm1.3)\!\times\!10^{-2}$ & $(4.1\pm1.8)\!\times\!10^{-2}$ & $(2.6\pm0.5)\!\times\!10^{0}$\\
 & Coverage & {\boldmath$0\pm0$} & {\boldmath$0\pm0$} & {\boldmath$0\pm0$} & {\boldmath$0\pm0$} & $\underline{(9.5\pm2.3)\!\times\!10^{-3}}$\\
\bottomrule
\end{tabular}
\end{table*}

For Table~\ref{tab:node_agreement}, we compute the between-node variance of each node's best commonly covered objective value and its fraction of commonly covered candidates using~\eqref{eq:experiment_node_variance}. These metrics assess agreement in solution quality and common-support satisfaction, respectively. Across all three tasks, \CPDST has the lowest mean quality variance and the lowest or tied-lowest coverage variance. Zero variance indicates equal node metrics, including when identity prediction with zero corrections preserves shared initial candidates. Together with Figs.~\ref{fig:beamforming}--\ref{fig:floris}, these results show strong worst-node performance with close node agreement. We next isolate calibration's contribution at equal traffic.

\subsection{Calibration across candidate counts and dimensions}
\label{subsec:calibration_validation}

We minimize the standard Rastrigin function~\cite{surjanovicRastrigin} on $[-5.12,5.12]^d$,
\begin{equation}
\label{eq:experiment_rastrigin_objective}
f_d(x)=10d+\sum_{j=1}^d\bigl[x_j^2-10\cos(2\pi x_j)\bigr],
\end{equation}
for $d\in\{6,12,24,48\}$. Each of four computing workers independently samples 4,096 input vectors $x$ uniformly on $[-5.12,5.12]^d$, evaluates~\eqref{eq:experiment_rastrigin_objective}, and stores $(x/5.12,-f_d(x)/4)$ as its offline dataset. We normalize inputs by 5.12 for training and sampling, then restore their original scale for evaluation. Thus nodes share the sampling distribution and $S=-f_d$, but hold different input--value pairs. All dimensions use $\beta=0.1$. We report the worst node's best objective value, $\max_i\min_q f_d(x_i^{(q)})$, over all raw candidates, and use coverage from~\eqref{eq:metrics} with $\mathcal C=[-5.12,5.12]^d$.

\begin{figure}[t]
\centering
\includegraphics[width=\columnwidth]{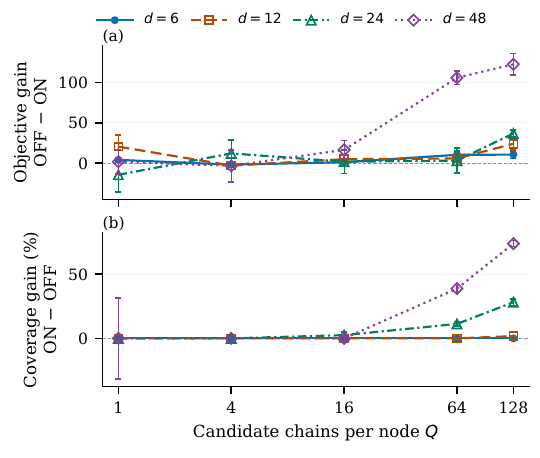}
\caption{Rastrigin at equal traffic of 12.375 MiB: (a) objective reduction (OFF minus ON) and (b) coverage increase (ON minus OFF, \%), with paired SE over five seeds. Positive values favor calibration.}
\label{fig:rastrigin}
\end{figure}

For $Q\in\{1,4,16,64,128\}$, we set the communication limit as 12,976,128 bytes with calibration enabled (ON) or disabled (OFF). Both use 16 levels, two prediction and correction-mixing rounds, and $\eta_\ell=0.04\sigma_{t_\ell}^2/n$. ON additionally spends 16 rounds per level on calibration, whereas OFF uses identity prediction and allocates the saved calibration traffic to extra corrections. Each policy distributes its correction steps as evenly as possible across levels, assigning any remainder to the final levels. ON and OFF share models and initial states, and use the same random noise for corresponding correction steps at each level. Figure~\ref{fig:rastrigin} reports the paired differences in objective and coverage.

Figure~\ref{fig:rastrigin} shows that calibration's gains in both objective quality and coverage become more pronounced at large $Q$, particularly as $d$ increases. A larger $Q$ spreads the shared calibration cost over more chains, while a larger $d$ makes each correction more expensive. Together, these trends highlight our algorithm's advantage in generating many high-dimensional candidates under a limited communication budget: calibrated prediction improves both solution quality and common-support satisfaction at the same communication cost.

\section{Conclusion}
\label{sec:conclusion}

We formulate decentralized offline optimization as sampling from a product-Gibbs target that combines local utilities on their common data support. Building on this formulation, \CPDST learns scores from historical records, predicts across noise levels through moment calibration, and corrects residual error through decentralized score tracking. Our analysis first controls each node's sampling error, then propagates it across levels and connects it to offline utility. We quantify the communication benefit of calibration through upper and lower bounds on correction counts, identifying when savings across $Q$ candidate chains outweigh the shared calibration cost. Experiments on beamforming, offloading, and wind-farm yaw control demonstrate improved solution quality and high coverage under tight communication budgets. Another experiment shows that calibration yields greater gains in both metrics when generating more candidate chains, particularly in higher dimensions.

\appendices
\section{Analytical Preliminaries}
\label{app:preliminaries}
\label{app:geometry_details}

In this part, we collect the score and coupling results used in Appendix~\ref{app:proofs}.

\subsection{Score identities and confinement}

\begin{proposition}[Product-score geometry]
\label{prop:harris_geometry}
We define $m_{i,t}(x)=\E[Z_i\mid\alpha_t Z_i+\sigma_t\xi=x]$, where $Z_i\sim q_{\beta,i}$ and $\xi\sim\mathcal N(0,I_d)$ are independent.  Under Assumption~\ref{ass:support},
\begin{align*}
 s_i(x,t)&=-x/\sigma_t^2+\alpha_t m_{i,t}(x)/\sigma_t^2,\\
 \nabla s_i(x,t)&=-I_d/\sigma_t^2+
 \alpha_t^2\Cov(Z_i\mid x)/\sigma_t^4.
\end{align*}
Thus $s_i(\cdot,t)$ is globally Lipschitz with bound $L_{i,t}=\sigma_t^{-2}+\alpha_t^2R_i^2\sigma_t^{-4}$.
Writing $R_\Sigma=\sum_iR_i$, the product drift satisfies
\begin{align*}
 \langle x,b_t(x)\rangle
 &\le-\frac n{\sigma_t^2}\|x\|^2+
 \frac{\alpha_tR_\Sigma}{\sigma_t^2}\|x\|,\\
 \langle x-y,b_t(x)-b_t(y)\rangle
 &\le-\frac n{\sigma_t^2}\|x-y\|^2+
 \frac{2\alpha_tR_\Sigma}{\sigma_t^2}\|x-y\|.
\end{align*}
In particular, for $\|x-y\|\ge4\alpha_tR_\Sigma/n$,
\begin{equation*}
 \langle x-y,b_t(x)-b_t(y)\rangle
 \le-\frac n{2\sigma_t^2}\|x-y\|^2.
\end{equation*}
\end{proposition}

\begin{proposition}[Learned-score confinement]
\label{prop:score_confinement}
If $g_{\theta_i}(\cdot,t)$ is $L_{g,i,t}$-Lipschitz, then $\operatorname{Lip}(\widehat s_i(\cdot,t))\le\sigma_t^{-2}+L_{g,i,t}$.
The learned sum $\sum_i\widehat s_i$ satisfies Proposition~\ref{prop:harris_geometry}'s dissipativity bounds. The local score error equals the residual error because the linear terms cancel.
\end{proposition}

\emph{Proof of Propositions~\ref{prop:harris_geometry} and~\ref{prop:score_confinement}.}

With noiseless labels, weighting by $e^{\beta f_i}$ transforms $p_i$ into $q_{\beta,i}$. For $X\sim q_{\beta,i}$ and $U_t\mid X\sim K_t(\cdot\mid X)$, the denoising score-matching identity~\cite[Eq.~(4.5) and Appendix]{vincent2011connection} gives
\begin{align}
 \E\|s_{\theta_i}(U_t,t)-\psi_t(U_t,X)\|^2
 ={}&\E\|s_{\theta_i}(U_t,t)-s_i(U_t,t)\|^2\notag\\
 &+\E\|\psi_t(U_t,X)-s_i(U_t,t)\|^2.
 \label{eq:score_loss_decomposition}
\end{align}
The last term is parameter-independent. Weighting by $\lambda(t)\sigma_t^2$ proves the loss identity. At $\alpha_t=0$, the architecture fixes $-u/\sigma_t^2$. Reward noise $\nu_i^{\rm obs}$ multiplies the Gibbs factor by $\E[e^{\beta\nu_i^{\rm obs}}\mid X=x]$, which cancels if constant in $x$.

Differentiating the Gaussian convolution gives the score identities. The bounds $\Cov(Z_i\mid x)\preceq R_i^2I_d$, $\|m_{i,t}(x)\|\le R_i$, and $\|m_{i,t}(x)-m_{i,t}(y)\|\le2R_i$ give Lipschitzness and dissipativity.

For $a_t=n/\sigma_t^2$, $G_t=\alpha_tR_\Sigma/\sigma_t^2$, and $V(x)=1+\|x\|^2$, the exact diffusion generator $\mathcal L_t=b_t\cdot\nabla+\Delta$, with $\Delta$ the Laplacian, satisfies
\begin{equation*}
 \mathcal L_tV(x)=2\langle x,b_t(x)\rangle+2d
 \le-a_t\|x\|^2+G_t^2/a_t+2d.
\end{equation*}
This verifies the Harris assumptions after rescaling by $1/\sqrt2$~\cite[Theorem~2.1 and Corollary~2.1]{eberle2019harris}. The diffusion $dZ_{t,s}=b_t(Z_{t,s})\,ds+\sqrt2\,dB_{t,s}$ has invariant density $\pi_t$. For $Z\sim\pi_t$, stationarity gives $\|Z\|_{L^2}\le M_{\pi,t}$ and $\|b_t(Z)\|_{L^2}\le B_{b,t}$, where
\begin{equation}
 M_{\pi,t}:=\sqrt{G_t^2/a_t^2+2d/a_t},\quad
 B_{b,t}:=a_tM_{\pi,t}+G_t.
 \label{eq:stationary_moment_bounds}
\end{equation}

The 1-Lipschitz output map gives $L_{g,i,t}\le\alpha_tR_i\operatorname{Lip}(F_{\theta_i}(\cdot,e(t)))/\sigma_t^2$. The bounded learned residual gives the same dissipativity estimates, and cancellation of the linear terms proves the error identity.

\subsection{Nonconvex Euler coupling}

\begin{lemma}[Nonconvex Euler coupling]
\label{thm:harris_corrector}
At fixed $t>0$, Proposition~\ref{prop:harris_geometry} yields an increasing, continuously differentiable $f_t$ and positive constants $c_t,\eta_{{\rm c},t},\kappa_{-,t},\kappa_{+,t},C_{f,t}$ such that
\begin{equation}
 \begin{aligned}
 \kappa_{-,t}(r+r^2)&\le f_t(r)\le\kappa_{+,t}(r+r^2),\\
 f_t'(r)&\le C_{f,t}(1+r).
 \end{aligned}
 \label{eq:cost_comparison}
\end{equation}
For exact-drift Euler steps $U',V'$ from $x,z$, a truncated reflection/mirror coupling gives
\begin{equation}
 \E f_t(\|U'-V'\|)
 \le(1-c_t\eta)f_t(\|x-z\|),
 \label{eq:euler_cost_contraction}
\end{equation}
for $0<\eta\le\eta_{{\rm c},t}$ chosen with $c_t\eta<1$.  With $L_{b,t}=\sum_i L_{i,t}$ and a constant $C_{0,t}$ uniform over this interval, its coupled distance obeys $\|U'-V'\|\le(1+\eta L_{b,t})\|x-z\|+C_{0,t}$.
\end{lemma}

\begin{proof}

We apply~\cite[Theorem~2.1, Lemma~2.3, and Theorem~2.5]{majka2020nonconvex} to the rescaled drift $u\mapsto b_t(\sqrt2u)/\sqrt2$. Proposition~\ref{prop:harris_geometry} verifies their Lipschitz and contractivity-at-infinity assumptions with
Lipschitz constant $L_{b,t}$, contractivity constant $a_t/2$, and radius $\mathcal R_t=\max\{1,2\sqrt2\alpha_tR_\Sigma/n\}$.
Their cost $f_t(r)=f(r/\sqrt2)$ and bounded reflected displacement yield~\eqref{eq:cost_comparison}--\eqref{eq:euler_cost_contraction} and the distance bound. For uniform constants, we take half their step threshold, with $c_0$ defined before their Eq.~(2.9):
\begin{equation}
 \begin{split}
 \eta_{{\rm c},t}=\frac12\min\left\{
 \frac{a_t}{2L_{b,t}^2},\frac8{a_t},\frac1{2L_{b,t}},\frac{\mathcal R_t^2}{4},\right.\\
 \left.\frac{2c_0\log(3/2)}{27L_{b,t}^2\mathcal R_t^2},
 \frac{c_0^2(\log2)^2}{144L_{b,t}^2\mathcal R_t^2}\right\}.
 \end{split}
 \label{eq:explicit_coupling_threshold}
\end{equation}
The cost $f$ and rate $c_t$ are specified by~\cite[Eqs.~(2.11), (2.13)]{majka2020nonconvex}, with $h_0=2\eta_{{\rm c},t}$ and $(L,K,R)=(L_{b,t},a_t/2,\mathcal R_t)$. These bounds are uniform in $\beta$ at fixed noise and prove the lemma.
\end{proof}
\subsection{Transport cost and step-size constants}
With $L_{g,i,t}$ denoting a Lipschitz bound for $g_{\theta_i}(\cdot,t)$, we define
\begin{align*}
 \widehat L_\ell&=\max_i\big(\sigma_{t_\ell}^{-2}+L_{g,i,t_\ell}\big),\\
 L_{s,\ell}&=\max_i\big(\sigma_{t_\ell}^{-2}
                  +\alpha_{t_\ell}^2R_i^2\sigma_{t_\ell}^{-4}\big).
\end{align*}
We set $\eta_{{\rm net},\ell}=(1-\gamma_\ell)^2/(2n\widehat L_\ell)$.
Lemma~\ref{thm:harris_corrector} supplies the cost-contraction rate for exact-drift Euler steps $c_t$ and the admissible step-size bound $\eta_{{\rm c},t}$ in~\eqref{eq:explicit_coupling_threshold}.
For a fixed $0<\kappa<1$, we use the joint step condition
\begin{equation}
 0<\eta_\ell\le\min\left\{1,\kappa\eta_{{\rm net},\ell},
                  \eta_{{\rm c},t_\ell},\frac{1}{2c_{t_\ell}}\right\}.
 \label{eq:joint_step}
\end{equation}

The lower bound in~\eqref{eq:cost_comparison} directly gives
$W_2^2(\Law(\bar x_\ell^k),\pi_{t_\ell})\le E_{\ell,k}/\kappa_{-,t_\ell}$,
using the single transport-cost definition~\eqref{eq:level_transport_cost}.

\section{Proof of Theorem~\ref{thm:fixed_corrector}(1)}
\label{app:proofs}
\label{app:proof_part1}
\label{app:tracking_details}
\label{app:adapted_coupling}

We use Appendix~\ref{app:preliminaries}'s constants and~\eqref{eq:joint_step}, and set $\delta_{\ell,k}=\sqrt nL_{s,\ell}r_{x,\ell}^k+\epsilon_{\Sigma,t_\ell}$.

\noindent\textbf{Step 1: Average invariant and moments.} The tracker invariant controls the average drift and its moments.

\begin{lemma}[Average invariant and moment envelope]
\label{lem:average_invariant}
For score fields~\eqref{eq:bounded_score_residual}, consistent initialization and doubly stochastic mixing give
$n\bar y_\ell^k=\sum_i\widehat s_i(x_{i,\ell}^k,t_\ell)$.
For $a_\ell=n/\sigma_{t_\ell}^2$, $G_\ell=\alpha_{t_\ell}R_\Sigma/\sigma_{t_\ell}^2$, condition~\eqref{eq:joint_step} gives $0<a_\ell\eta_\ell<1/2$ and
\begin{align*}
 \sup_k\|\bar x_\ell^k\|_{L^2}
 &\le M_{x,\ell}:=\|\bar x_\ell^0\|_{L^2}+G_\ell/a_\ell
 +\sqrt{\frac{4d}{3a_\ell}},\\
 \sup_k\|n\bar y_\ell^k\|_{L^2}&\le M_{b,\ell}:=a_\ell M_{x,\ell}+G_\ell.
\end{align*}
\end{lemma}

\emph{Proof.}

Double stochasticity telescopes the tracker innovations, proving the invariant and its preservation under refresh. By~\eqref{eq:bounded_score_residual},
\begin{equation*}
 \bar x_\ell^{k+1}=(1-a_\ell\eta_\ell)\bar x_\ell^k
 +\eta_\ell{\textstyle\sum_i}g_{\theta_i}(x_{i,\ell}^k,t_\ell)+\sqrt{2\eta_\ell}\xi_{\ell,k}.
\end{equation*}
Unrolling bounds the residual sum by $G_\ell/a_\ell$ and the Gaussian sum's second moment by $2d/[a_\ell(2-a_\ell\eta_\ell)]$. Minkowski gives $M_{x,\ell}$, and the invariant gives $M_{b,\ell}$.

\noindent\textbf{Step 2: State and tracker disagreement.} Step 1 bounds the score innovations in the following recursion.

\begin{lemma}[Closed tracking bound]
\label{thm:fixed_tracking}
For $0<\kappa<1$ and $\eta_{{\rm net},\ell}$ defined before~\eqref{eq:joint_step}, we assume $0<\eta_\ell\le\kappa\eta_{{\rm net},\ell}$.
For this recursion, we write $u_\ell^k=(r_{x,\ell}^k,r_{y,\ell}^k)^\top$ and define
\begin{align*}
 H_\ell&=\begin{bmatrix}
 \gamma_\ell&\eta_\ell n\\
 \widehat L_\ell(1+\gamma_\ell)&\gamma_\ell+\eta_\ell n\widehat L_\ell
 \end{bmatrix},\\
 G_\ell^{\rm trk}&=\widehat L_\ell\sqrt n\,(\eta_\ell M_{b,\ell}+\sqrt{2\eta_\ell d}).
\end{align*}
Writing $\varrho(H_\ell)$ for the spectral radius and $\preceq$ for componentwise inequality, we have $\varrho(H_\ell)<1$ and
\begin{equation}
 u_\ell^k\preceq H_\ell^k u_\ell^0+
 \sum_{j=0}^{k-1}H_\ell^j(0,G_\ell^{\rm trk})^\top.
 \label{eq:comparison_solution}
\end{equation}
For $\Delta_{H,\ell}=(1-\gamma_\ell)^2-2\eta_\ell n\widehat L_\ell>0$,
\begin{equation*}
 \limsup_k r_{x,\ell}^k\le\frac{\eta_\ell nG_\ell^{\rm trk}}{\Delta_{H,\ell}},\quad
 \limsup_k r_{y,\ell}^k\le\frac{(1-\gamma_\ell)G_\ell^{\rm trk}}{\Delta_{H,\ell}}.
\end{equation*}
In addition,
\begin{equation*}
 \|ny_{i,\ell}^k-b_{t_\ell}(\bar x_\ell^k)\|_{L^2}
 \le nr_{y,\ell}^k+\sqrt nL_{s,\ell} r_{x,\ell}^k+\epsilon_{\Sigma,t_\ell}.
\end{equation*}
\end{lemma}

At fixed geometry, these upper bounds on long-run disagreement are $O(\eta_\ell^{3/2})$ and $O(\sqrt{\eta_\ell})$, respectively.

\begin{corollary}[Block-mixed time-varying networks]
\label{thm:time_tracking}
Under Assumption~\ref{ass:graph}, Lemma~\ref{thm:fixed_tracking} applies with $\gamma_\ell=\rho_B^{\lfloor m_\ell/B\rfloor}<1$ for any integer $m_\ell\ge B$.
\end{corollary}

\emph{Proof of Lemma~\ref{thm:fixed_tracking} and Corollary~\ref{thm:time_tracking}.}

Synchronized noise vanishes under $\Pi$. Using $G_\ell^{\rm trk}$, projection and the full increment give
\begin{align*}
 r_{x,\ell}^{k+1}&\le\gamma_\ell r_{x,\ell}^k+\eta_\ell n r_{y,\ell}^k,\\
 \|X_\ell^{k+1}-X_\ell^k\|_{L^2}
 &\le(1+\gamma_\ell)r_{x,\ell}^k+\eta_\ell n r_{y,\ell}^k+G_\ell^{\rm trk}/\widehat L_\ell.
\end{align*}
The $\widehat L_\ell$-Lipschitz score innovations therefore yield
$u_\ell^{k+1}\preceq H_\ell u_\ell^k+(0,G_\ell^{\rm trk})^\top$.
The step condition makes $I-H_\ell$ a nonsingular $M$-matrix with determinant $\Delta_{H,\ell}>0$. Iteration proves~\eqref{eq:comparison_solution}, and multiplication by $(I-H_\ell)^{-1}$ gives both floors. Adding and subtracting $\sum_i s_i(x_{i,\ell}^k,t_\ell)$ proves the drift-error bound. Since $\widehat L_\ell\ge\sigma_{t_\ell}^{-2}$ implies $a_\ell\eta_\ell\le\kappa/2$, the average and disagreement estimates close the moment bounds.

The block-product bound in Section~\ref{subsec:communication_graph} proves the corollary.

\noindent\textbf{Step 3: Perturbed transport contraction.} Steps 1--2 express the average as a perturbed Euler step:

\begin{equation*}
 \bar x_\ell^{k+1}=\bar x_\ell^k+\eta_\ell b_{t_\ell}(\bar x_\ell^k)
 +\sqrt{2\eta_\ell}\xi_{\ell,k}+\eta_\ell \mathfrak e_{\ell,k}.
\end{equation*}
Here $\mathfrak e_{\ell,k}=\sum_i[\widehat s_i(x_{i,\ell}^k,t_\ell)-s_i(\bar x_\ell^k,t_\ell)]$ is pre-update measurable and $\|\mathfrak e_{\ell,k}\|_{L^2}\le\delta_{\ell,k}$.

We couple $\bar x_\ell^0$ to stationary $Z_{\ell,0}\sim\pi_{t_\ell}$ with cost approaching $E_{\ell,0}$, conditioning the network on $\bar x_\ell^0$. We then evolve the reference diffusion. Given the full past, Lemma~\ref{thm:harris_corrector} couples Euler steps $\mathsf U_{\ell,k+1}$ from $\bar x_\ell^k$ and $\mathsf V_{\ell,k+1}$ from $Z_{\ell,k\eta_\ell}$, using the reference Brownian increment for the latter. The former's conditionally Gaussian noise, shared across nodes, preserves the algorithm's joint law.

We write
\[
 \mathfrak r_{\ell,k}=\int_{k\eta_\ell}^{(k+1)\eta_\ell}
 [b_{t_\ell}(Z_{\ell,s})-b_{t_\ell}(Z_{\ell,k\eta_\ell})]\,ds .
\]
The actual next state and reference are $\mathsf U_{\ell,k+1}+\eta_\ell\mathfrak e_{\ell,k}$ and $\mathsf V_{\ell,k+1}+\mathfrak r_{\ell,k}$. The local Euler argument in~\cite[Lemma~2.14]{majka2020nonconvex}, applied to the stationary diffusion above, gives for $\eta_\ell\le1$
\begin{equation*}
 \|\mathfrak r_{\ell,k}\|_{L^2}\le L_{0,\ell}\eta_\ell^{3/2},\quad
 L_{0,\ell}=L_{b,t_\ell}\sqrt{2B_{b,t_\ell}^2/3+2d}.
\end{equation*}

For $\mathfrak d_{\ell,k}=\|\mathsf U_{\ell,k+1}-\mathsf V_{\ell,k+1}\|$ and $\mathfrak s_{\ell,k}=\|\eta_\ell \mathfrak e_{\ell,k}-\mathfrak r_{\ell,k}\|$, monotonicity and the derivative bound give
\[
 \begin{aligned}
 &f_{t_\ell}(\|\bar x_\ell^{k+1}-Z_{\ell,(k+1)\eta_\ell}\|)\\
 &\quad\le f_{t_\ell}(\mathfrak d_{\ell,k})
       +C_{f,t_\ell}(1+\mathfrak d_{\ell,k})\mathfrak s_{\ell,k}
       +C_{f,t_\ell}\mathfrak s_{\ell,k}^2.
 \end{aligned}
\]
The coupled-distance bound implies
\[
 \|1+\mathfrak d_{\ell,k}\|_{L^2}\le
 M_{*,\ell}:=1+C_{0,t_\ell}+(1+L_{b,t_\ell})(M_{x,\ell}+M_{\pi,t_\ell}).
\]
Consequently, for $D_{\ell,k}=\E f_{t_\ell}(\|\bar x_\ell^k-Z_{\ell,k\eta_\ell}\|)$,
\begin{equation*}
 D_{\ell,k+1}\le q_\ell D_{\ell,k}+C_{1,\ell}\eta_\ell\delta_{\ell,k}
 +C_{2,\ell}\eta_\ell^2\delta_{\ell,k}^2+C_{3,\ell}\eta_\ell^{3/2}.
\end{equation*}
with the admissible choices
\begin{align*}
 C_{1,\ell}&=C_{f,t_\ell}M_{*,\ell},\qquad C_{2,\ell}=2C_{f,t_\ell},\\
 C_{3,\ell}&=C_{f,t_\ell}M_{*,\ell}L_{0,\ell}+2C_{f,t_\ell}L_{0,\ell}^2 .
\end{align*}
These follow from $\|\mathfrak s_{\ell,k}\|_{L^2}\le\eta_\ell\delta_{\ell,k}+L_{0,\ell}\eta_\ell^{3/2}$, $\eta_\ell\le1$, and unconditional Cauchy--Schwarz.

\noindent\textbf{Step 4: Accumulated error.} Step 3's constants are uniform at fixed geometry and initial-moment bounds. Summing its recursion gives
\begin{align}
 \Xi_\ell(R_\ell)={}&\frac{C_{3,\ell}}{c_{t_\ell}}(1-q_\ell^{R_\ell})\sqrt{\eta_\ell}\notag\\
 &+\eta_\ell\sum_{k=0}^{R_\ell-1}q_\ell^{R_\ell-1-k}
 (C_{1,\ell}\delta_{\ell,k}+C_{2,\ell}\eta_\ell\delta_{\ell,k}^2).
 \label{eq:corrector_cost_floor}
\end{align}
Iteration and an initial coupling approaching its infimum give

\begin{equation}
 E_{\ell,R_\ell}
 \le q_\ell^{R_\ell}E_{\ell,0}+\Xi_\ell(R_\ell).
 \label{eq:average_cost_bound}
\end{equation}

\noindent\textbf{Step 5: Explicit error envelope.} To substitute Step 2 into Step 4, we set $d_\ell^{\rm trk}=2\widehat L_\ell/(1-\gamma_\ell)$,
$\theta_\ell=1-(1-\kappa)(1-\gamma_\ell)^2/2$, and
$T_\ell^{\rm trk}=\max\{r_{x,\ell}^0,r_{y,\ell}^0/d_\ell^{\rm trk}\}$.
Since $H_\ell(1,d_\ell^{\rm trk})^\top\preceq\theta_\ell(1,d_\ell^{\rm trk})^\top$,~\eqref{eq:comparison_solution} gives
\begin{equation}
 r_{x,\ell}^k\le T_\ell^{\rm trk}\theta_\ell^k+F_\ell,\qquad
 F_\ell=\frac{\eta_\ell nG_\ell^{\rm trk}}{\Delta_{H,\ell}}.
 \label{eq:explicit_tracking_envelope}
\end{equation}
We set $D_\ell=\sqrt nL_{s,\ell}T_\ell^{\rm trk}$ and $e_\ell=\epsilon_{\Sigma,t_\ell}+\sqrt nL_{s,\ell}F_\ell$.
For $R\ge1$, we define $\mathcal S_R(q,z)=(q^R-z^R)/(q-z)$ if $q\ne z$, and $\mathcal S_R(q,q)=Rq^{R-1}$.  We set $\mathcal S_0=0$.
Substituting $\delta_{\ell,k}\le D_\ell\theta_\ell^k+e_\ell$ into~\eqref{eq:corrector_cost_floor} yields
\begin{align}
 \Xi_\ell(R_\ell)\le{}&\frac{C_{3,\ell}\sqrt{\eta_\ell}+C_{1,\ell}e_\ell
                  +2C_{2,\ell}\eta_\ell e_\ell^2}{c_{t_\ell}}\notag\\
 &+\eta_\ell D_\ell\big[C_{1,\ell}\mathcal S_{R_\ell}(q_\ell,\theta_\ell)
 +2C_{2,\ell}\eta_\ell D_\ell\mathcal S_{R_\ell}(q_\ell,\theta_\ell^2)\big].
 \label{eq:explicit_cost_envelope}
\end{align}
Envelopes~\eqref{eq:explicit_tracking_envelope}--\eqref{eq:explicit_cost_envelope} evaluate~\eqref{eq:round_requirement}. Since $F_\ell=O(\eta_\ell^{3/2})$ and $\theta_\ell<1$ uniformly at fixed geometry, Remark~\ref{rem:vanishing_correction}'s joint limits remove every term. The Euler term contributes $O(\eta_\ell^{1/4})$ to the $W_2$ upper bound.

Finally, coupling each node to the network average and using~\eqref{eq:cost_comparison} gives
\[
 W_2(\Law(x_{i,\ell}^{R_\ell}),\pi_{t_\ell})
 \le r_{x,\ell}^{R_\ell}+\sqrt{E_{\ell,R_\ell}/\kappa_{-,t_\ell}}.
\]
Substituting~\eqref{eq:average_cost_bound} proves~\eqref{eq:fixed_corrector_bound}, completing Theorem~\ref{thm:fixed_corrector}(1).

\section{Proof of Theorem~\ref{thm:fixed_corrector}(2)}
\label{app:proof_part2}

\noindent\textbf{Step 1: Tracking upper bound.} With diffusion noise level, models, mixing bound, and initial moment bounds fixed, $G_\ell^{\rm trk}=O(\sqrt{\eta_\ell})$ and $\Delta_{H,\ell}\to(1-\gamma_\ell)^2>0$. The first component of~\eqref{eq:comparison_solution} therefore gives
\[
 \limsup_{k\to\infty}r_{x,\ell}^k
 \le\frac{\eta_\ell nG_\ell^{\rm trk}}{\Delta_{H,\ell}}
 =O(\eta_\ell^{3/2}).
\]

\noindent\textbf{Step 2: Local-score lower bound.} For the baseline replacing the tracker by local scores, we write $Z_\ell^k=\col(z_{i,\ell}^k)_{i=1}^n$, $\bar z_\ell^k=n^{-1}\sum_i z_{i,\ell}^k$, and $a_{\ell,k}=\|\Pi Z_\ell^k\|_{L^2}$. The local-score heterogeneity condition is
\begin{equation}
 h_{\ell,k}:=\|\Pi\widehat S_\ell(\one\otimes\bar z_\ell^k)\|_{L^2},
 \quad\limsup_k h_{\ell,k}\ge h_{*,\ell}>0,
 \label{eq:persistent_heterogeneity}
\end{equation}
with $h_{*,\ell}$ independent of sufficiently small $\eta_\ell$. Shared noise cancels under $\Pi$, yielding
\[
 \Pi Z_\ell^{k+1}=P_{\ell,k}\Pi Z_\ell^k
 +\eta_\ell n\Pi\widehat S_\ell(Z_\ell^k).
\]
Adding and subtracting $\widehat S_\ell(\one\otimes\bar z_\ell^k)$, then using Lipschitz continuity and the triangle inequality, gives
\begin{align*}
 \eta_\ell n h_{\ell,k}&\le a_{\ell,k+1}
 +(\gamma_\ell+\eta_\ell n\widehat L_\ell)a_{\ell,k},\\
 \limsup_k a_{\ell,k}&\ge
 \frac{\eta_\ell n\limsup_k h_{\ell,k}}
 {1+\gamma_\ell+\eta_\ell n\widehat L_\ell}\\
 &\ge\frac{\eta_\ell n h_{*,\ell}}
 {1+\gamma_\ell+\eta_\ell n\widehat L_\ell}
 =\Omega(\eta_\ell).
\end{align*}
Together with Step 1, this proves Theorem~\ref{thm:fixed_corrector}(2). Condition~\eqref{eq:persistent_heterogeneity} is used only in Step 2. With identical local score fields and a shared initial state, the baseline preserves equal node states and $a_{\ell,k}=0$ for all $k$.

\section{Proof of Theorem~\ref{thm:end_to_end}}
\label{app:finite_consensus}

We write $z_{i,\ell}^{\rm tr}=(P_\ell^{\rm tr}X_{\ell-1}^{R_{\ell-1}})_i$ and $v_{i,\ell}=\Phi_{i,\ell}(z_{i,\ell}^{\rm tr})-\Phi_\ell(z_{i,\ell}^{\rm tr})$. We set $e_{v,\ell}=\|\col(v_{i,\ell})_{i=1}^n\|_{L^2}/\sqrt n$. Affine averaging and projection give
\begin{equation}
 \begin{aligned}
 \|\bar x_\ell^0-\Phi_\ell(\bar x_{\ell-1}^{R_{\ell-1}})\|_{L^2}
 &\le e_{v,\ell},\\
 r_{x,\ell}^0&\le K_\ell\gamma_\ell^{\rm tr}r_{x,\ell-1}^{R_{\ell-1}}+\sqrt n\,e_{v,\ell}.
 \end{aligned}
 \label{eq:finite_calibration_consensus}
\end{equation}
For node-specific maps, we use the adjusted coefficients
\begin{equation}
 \begin{aligned}
 \mathcal A_\ell&=\frac{3\kappa_{+,t_\ell}}{\kappa_{-,t_{\ell-1}}}
                  \max(K_\ell,K_\ell^2),\\
 \mathcal B_\ell&=3\kappa_{+,t_\ell}
                  (b_\ell+b_\ell^2+e_{v,\ell}+e_{v,\ell}^2).
 \end{aligned}
 \label{eq:finite_consensus_coefficients}
\end{equation}
Here $e_{v,\ell}$ measures errors at visited inputs, whereas~\eqref{eq:calibration_estimation_error} uses target-distributed inputs.

We couple $U_{\ell-1}\sim\Law(\bar x_{\ell-1}^{R_{\ell-1}})$ to $Z_{\ell-1}\sim\pi_{t_{\ell-1}}$ at cost approaching $E_{\ell-1,R_{\ell-1}}$, then glue an optimal coupling of $\Phi_\ell(Z_{\ell-1})$ and $V_\ell\sim\pi_{t_\ell}$. For a common map, $\bar x_\ell^0=\Phi_\ell(U_{\ell-1})$ and
\[
 \|\Phi_\ell(U_{\ell-1})-V_\ell\|
 \le K_\ell\|U_{\ell-1}-Z_{\ell-1}\|
       +\|\Phi_\ell(Z_{\ell-1})-V_\ell\|.
\]
For $h(r)=r+r^2$, we use $h(r+s)\le2h(r)+2h(s)$, $h(Kr)\le\max(K,K^2)h(r)$, and~\eqref{eq:cost_comparison} to obtain $E_{\ell,0}\le\mathcal A_\ell E_{\ell-1,R_{\ell-1}}+\mathcal B_\ell$. Substitution into~\eqref{eq:average_cost_bound} gives the one-level recursion in Theorem~\ref{thm:end_to_end}.

For node-specific maps, we add the average map error $n^{-1}\sum_i v_{i,\ell}$. Its $L^2$ bound $e_{v,\ell}$ and $h(r+s+u)\le3[h(r)+h(s)+h(u)]$ give~\eqref{eq:finite_consensus_coefficients}. Tracker refresh gives
\begin{align*}
 r_{y,\ell}^0\le{}&\gamma_\ell^{\rm tr}r_{y,\ell-1}^{R_{\ell-1}}\\
 &+\big\|\col\big(\widehat s_i(x_{i,\ell}^0,t_\ell)
       -\widehat s_i(x_{i,\ell-1}^{R_{\ell-1}},t_{\ell-1})\big)_{i=1}^n\big\|_{L^2}.
\end{align*}
Affine growth and Lipschitz scores ensure finite moments.  We use this refreshed initial disagreement in~\eqref{eq:comparison_solution}.

Thus the same recursion holds for node-specific maps with~\eqref{eq:finite_consensus_coefficients}. Iterating over $\ell=1,\ldots,L$ gives~\eqref{eq:end_to_end}. Since $t_L=\varepsilon$, the cost comparison and the node-to-average coupling yield
\[
 W_2(\Law(x_{i,L}^{R_L}),\pi_\varepsilon)
 \le r_{x,L}^{R_L}+\sqrt{E_{L,R_L}/\kappa_{-,t_L}}.
\]
The inherited-cost coefficient is $q_\ell^{R_\ell}\mathcal A_\ell$, so the inherited-error term in the bound contracts when this product is below one. This proves Theorem~\ref{thm:end_to_end}.

\section{Proof of Theorem~\ref{thm:optimality}}
\label{app:offline_value}

\noindent\textbf{Step 1: Value bound.} Coverage~\eqref{eq:optimum_coverage} makes the denominator below positive. Restricting the Gibbs normalizer to $U_{\delta/4}$ gives
\[
 q_{\beta,G}\{S\le S_\cap^\star-\delta/2\}
 \le e^{-\beta\delta/4}/p_\cap(U_{\delta/4}).
\]
For a $W_2$-optimal coupling $(X,Y)$ with $X\sim\mu_i$ and $Y\sim q_{\beta,G}$, a value gap $\delta$ at $X$ implies a gap $\delta/2$ at $Y$ or $\|X-Y\|>\delta/(2L_S)$. Markov's inequality bounds the latter probability by $4L_S^2\Delta_i^2/\delta^2$, proving~\eqref{eq:sampling_to_value}. If $L_S=0$, the value gap is zero. Substitution of~\eqref{eq:optimality_wasserstein_budget} gives failure probability at most $\nu$.

\noindent\textbf{Step 2: Terminal sampling error.} Theorem~\ref{thm:end_to_end} and the triangle inequality give the stated $\Delta_i$. For bounded compactly supported factors, dilation continuity and Gaussian approximate identities give $q_{\beta,i,t}\to q_{\beta,i}$ in $L^1$ as $\alpha_t\to1$ and $\sigma_t\to0$. Uniform $L^\infty$ bounds, a telescoping product expansion, and the positive limiting normalizer give $\pi_t\to q_{\beta,G}$ in $L^1$. The bound $\pi_t\le Cq_{\beta,1,t}$ gives uniform fourth moments, hence $\delta_{\rm sm}(t)\to0$ at fixed $\beta$.

\noindent\textbf{Step 3: Feasible projection.} For a measurable nearest-point projection onto $K=\mathcal X_\cap$, $Y\in K$ almost surely and $\|T_K(X)-Y\|\le2\|X-Y\|$. Thus the projected law is feasible and has $W_2$ error at most $2\Delta_i$. For convex $K$, nonexpansiveness improves this to $\Delta_i$. Applying Step 1 with these errors proves the projected-output bound and completes Theorem~\ref{thm:optimality}.

\section{Proof of Theorem~\ref{thm:predictor_separation}}
\label{app:prediction}

We use the preceding level's accuracy requirement, or the initialization assumption for $\ell=1$, as the incoming network-average error bound $e_{\ell,0}=\tau_{\ell-1}$ for every chain. Steps 1--3 derive the sufficient calibrated budget~\eqref{eq:predictor_tolerance_test}, and Step 4 derives the necessary uncalibrated budget~\eqref{eq:full_run_lower_counts}.

\noindent\textbf{Step 1: Ideal prediction error.} We denote by $(m_j^\star,C_j^\star)$ the Gaussian-product moments in~\eqref{eq:product_gaussian} computed from exact local moments. We set $g_j=\mathcal N(m_j^\star,C_j^\star)$, $A_\ell^\star=(C_\ell^\star)^{1/2}(C_{\ell-1}^\star)^{-1/2}$, and $\Phi_\ell^\star(x)=m_\ell^\star+A_\ell^\star(x-m_{\ell-1}^\star)$. Since $(\Phi_\ell^\star)_\#g_{\ell-1}=g_\ell$, the triangle inequality and Lipschitz pushforward stability bound the target mismatch by
\begin{equation}
 \epsilon_j^G=W_2(\pi_{t_j},g_j),\qquad
 B_\ell^G=\|A_\ell^\star\|_{\rm op}\epsilon_{\ell-1}^G+\epsilon_\ell^G.
 \label{eq:gaussian_transport_proxy}
\end{equation}

Using the center and radii from Theorem~\ref{thm:predictor_separation}, we write $Z_i=c+U_i\sim q_{\beta,i}$, set $\nu_i=\Law(U_i)$, and define
\begin{align*}
 \bar U&=n^{-1}\sum_iU_i,\quad
 V(U)=\sum_i\|U_i-\bar U\|^2,\\
 \lambda_t&=\alpha_t^2/(2\sigma_t^2),\quad
 dP_t(U)\propto e^{-\lambda_t V(U)}\prod_i\nu_i(dU_i).
\end{align*}
For $U_t=(U_{1,t},\ldots,U_{n,t})\sim P_t$, we set $\bar U_t=n^{-1}\sum_iU_{i,t}$ and take independent $G\sim\mathcal N(0,I_d)$.  Completing the square gives
\begin{equation*}
 \pi_t=\Law\left(\alpha_tc+\alpha_t\bar U_t+
             \frac{\sigma_t}{\sqrt n}G\right),
\end{equation*}

We use $\|\cdot\|_{\rm op}$ and $\|\cdot\|_{\rm F}$ for operator and Frobenius norms. We set $d_j=m_j^\star-\alpha_{t_j}c$ for $j\in\{\ell-1,\ell\}$.  If $\delta_{{\rm lat},\ell}$ bounds the total variation $\operatorname{TV}(P_{t_{\ell-1}},P_{t_\ell})$, the ideal mismatch is also at most $B_\ell^{\rm tr}$, where
\begin{align}
 B_\ell^{\rm tr}:={}&\|d_\ell-A_\ell^\star d_{\ell-1}\|
 +r_{\rm av}\|\alpha_{t_{\ell-1}}A_\ell^\star-\alpha_{t_\ell}I_d\|_{\rm op}\notag\\
 &+2\alpha_{t_\ell}r_{\rm av}\sqrt{\delta_{{\rm lat},\ell}}
 +\frac{\|\sigma_{t_{\ell-1}}A_\ell^\star-\sigma_{t_\ell}I_d\|_{\rm F}}{\sqrt n}.
 \label{eq:latent_transport_bound}
\end{align}
A distribution-free choice is
\begin{equation*}
 \delta_{{\rm lat},\ell}
 =\min\left\{1,\frac{|\lambda_{t_\ell}-\lambda_{t_{\ell-1}}|}{4}
                      \sum_i r_i^2\right\}.
\end{equation*}

We use a maximal coupling of the latent laws and share $G$. Minkowski and $\|\bar U_{t_{\ell-1}}-\bar U_{t_\ell}\|_{L^2}\le2r_{\rm av}\sqrt{\delta_{{\rm lat},\ell}}$ give the four terms in~\eqref{eq:latent_transport_bound}.
For the tilted law $P_\lambda$, boundedness permits differentiation and yields
\[
 \frac12\int|\partial_\lambda dP_\lambda|
 =\frac12\E_{P_\lambda}|V-\E_{P_\lambda}V|
 \le\frac{\sup V-\inf V}{4}.
\]
Since $0\le V=\sum_i\|U_i\|^2-n\|\bar U\|^2\le\sum_i r_i^2$, integration gives the stated total-variation bound, with $\operatorname{TV}(P,Q)=\sup_E|P(E)-Q(E)|$.

Combining~\eqref{eq:gaussian_transport_proxy} and~\eqref{eq:latent_transport_bound}, we may use $B_\ell=\min\{B_\ell^G,B_\ell^{\rm tr}\}$ for the ideal predictor.

\noindent\textbf{Step 2: Calibration and network errors.} Step 1 uses exact moments. The calibration error is
\begin{equation}
 e_\ell^{\rm cal}=\|\widehat\Phi_\ell(Z)-\Phi_\ell^\star(Z)\|_{L^2},
 \qquad Z\sim\pi_{t_{\ell-1}}.
 \label{eq:calibration_estimation_error}
\end{equation}
Conditional on frozen models, we assume each node's $M$ projected samples are independent of each other and inference, with law $q_i^{\rm gen}$ satisfying $\max_iW_2(q_i^{\rm gen},q_{\beta,i})\le\chi$. Under exact aggregation, a computable $K_{\rm cal}>0$, independent of $M,\chi,\zeta$, gives
\begin{equation}
 e_\ell^{\rm cal}\le K_{\rm cal}\left[
 \chi+\sqrt{\frac{\log(4nd/\zeta)}{2M}}\right]
 \label{eq:calibration_sample_bound}
\end{equation}
with probability at least $1-\zeta$, $0<\zeta<1$, simultaneously over a fixed positive-noise schedule.

To establish this bound, we set $R_{\max}=\max_iR_i$,
$v_{\min}=\min_{0\le j\le L}\sigma_{t_j}^2>0$,
$v_{\max}=\max_{0\le j\le L}(\sigma_{t_j}^2+\alpha_{t_j}^2R_{\max}^2)$. For $u_M=\sqrt{\log(4nd/\zeta)/(2M)}$, Hoeffding and a union bound over $2nd$ moments bound the mean/variance errors by $2R_{\max}u_M$ and $5R_{\max}^2u_M$. A $W_2$ coupling adds bias bounds $\chi$ and $4R_{\max}\chi$, yielding
\[
 e_\mu=\bar\alpha(\chi+2R_{\max}u_M),\qquad
 e_D=\bar\alpha^2(4R_{\max}\chi+5R_{\max}^2u_M).
\]
We set $K_*=\sqrt{v_{\max}/v_{\min}}$. Local/product variances remain in $[v_{\min},v_{\max}]$/$[v_{\min}/n,v_{\max}/n]$. Differentiating~\eqref{eq:product_gaussian}--\eqref{eq:calibrated_map} gives
\begin{align*}
 |\widehat m_{\ell,j}-m_{\ell,j}^\star|&\le e_m:=e_\mu+2\bar\alpha R_{\max}e_D/v_{\min},\\
 |\widehat C_{\ell,jj}-C_{\ell,jj}^\star|&\le v_{\max}e_D/(nv_{\min}),\\
 \|\widehat A_\ell-A_\ell^\star\|_{\rm op}&\le e_A:=\frac{v_{\max}(1+K_*)e_D}{2v_{\min}^2},\quad
 \|\widehat A_\ell\|_{\rm op}\le K_*.
\end{align*}
Using~\eqref{eq:stationary_moment_bounds} in~\eqref{eq:calibration_estimation_error},
\begin{equation*}
 e_\ell^{\rm cal}\le\sqrt d(1+K_*)e_m+
 (M_{\pi,t_{\ell-1}}+\sqrt d\,\bar\alpha R_{\max})e_A.
\end{equation*}
We take $K_{\rm cal}$ as the maximum of one and the coefficients of $\chi,u_M$ over transitions.

If each auxiliary local Langevin sampler satisfies Theorem~\ref{thm:fixed_corrector}'s moment, step-size, and score-error conditions along its own trajectory, the $n=1$ bound at $\varepsilon_{\rm g}>0$ gives $\Delta_i^{\rm loc}$. Nonexpansive ball projection and Gaussian corruption then give
\[
 \chi\le\max_i\{\Delta_i^{\rm loc}
       +|1-\alpha_{\varepsilon_{\rm g}}|R_i
       +\sigma_{\varepsilon_{\rm g}}\sqrt d\}.
\]
For a prescribed calibration-error tolerance $\tau_{\rm cal}>K_{\rm cal}\chi$, bound~\eqref{eq:calibration_sample_bound} is at most $\tau_{\rm cal}$ whenever
\begin{equation}
 M\ge\left\lceil\frac{\log(4nd/\zeta)}
 {2(\tau_{\rm cal}/K_{\rm cal}-\chi)^2}\right\rceil .
 \label{eq:calibration_batch_budget}
\end{equation}
For the implemented local maps, the finite-consensus error is
\[
 e_\ell^{\rm net}=\max_i
 \|\Phi_{i,\ell}(z_{i,\ell}^{\rm tr})-\widehat\Phi_\ell(z_{i,\ell}^{\rm tr})\|_{L^2},
 \quad z_{i,\ell}^{\rm tr}=(P_\ell^{\rm tr}X_{\ell-1}^{R_{\ell-1}})_i.
\]
If $\Phi_{i,\ell}(x)=a_{i,\ell}+A_{i,\ell}x$ and $\widehat\Phi_\ell(x)=\widehat a_\ell+\widehat A_\ell x$, the consensus error satisfies
$e_\ell^{\rm net}\le\max_i[\|a_{i,\ell}-\widehat a_\ell\|+
\|A_{i,\ell}-\widehat A_\ell\|_{\rm op}\|z_{i,\ell}^{\rm tr}\|_{L^2}]$.
Here $m_j^{\rm cal}$ counts calibration mixing rounds at endpoint $j$, and $H_j^{\rm cal}$ stacks its local diagonal precision entries and information vectors.  We write $r_j^{\rm cal}=\|(\Pi\otimes I_{2d})H_j^{\rm cal}\|_2$.  For $r_j^{\rm cal}>0$ and $0<\rho_B<1$, a tolerance $u>0$ for mixed averages (or $nu$ for reconstructed sums) is ensured by
\begin{equation}
 m_j^{\rm cal}\ge B\left\lceil
 \frac{[\log(r_j^{\rm cal}/u)]_+}{\log(1/\rho_B)}\right\rceil,
 \quad j\in\{\ell-1,\ell\}.
 \label{eq:calibration_consensus_budget}
\end{equation}
Zero initial disagreement needs no rounds, while $\rho_B=0$ needs at most $B$. Lipschitz continuity of inversion and square roots on the positive precision range converts statistic errors to map errors on the same calibration event.

\noindent\textbf{Step 3: Sufficient calibrated budget.} Combining Steps 1--2 with the incoming error bound gives $v_\ell$ from Theorem~\ref{thm:predictor_separation}. We choose the initial-cost bound
\begin{equation}
 D_{\ell,\rm in}^{\rm on}\ge\kappa_{+,t_\ell}(v_\ell+v_\ell^2),
 \qquad D_{\ell,\rm in}^{\rm on}>0.
 \label{eq:calibrated_initial_budget}
\end{equation}
Affine averaging,~\eqref{eq:cost_comparison}, and~\eqref{eq:calibrated_initial_budget} yield $E_{\ell,0}\le D_{\ell,\rm in}^{\rm on}$. For any $D_{\ell,\rm in}>0$ bounding $E_{\ell,0}$, equations~\eqref{eq:fixed_corrector_bound} and~\eqref{eq:average_cost_bound} ensure the prescribed $W_2$ tolerance for every node and the network average whenever
\begin{equation}
 \begin{aligned}
 r_{x,\ell}^{R_\ell}&\le\tau_\ell/2,\quad
 \Xi_\ell(R_\ell)\le\kappa_{-,t_\ell}\tau_\ell^2/8,\\
 R_\ell&\ge\left\lceil
 \frac{[\log(8D_{\ell,\rm in}/(\kappa_{-,t_\ell}\tau_\ell^2))]_+}
 {-\log q_\ell}\right\rceil.
 \end{aligned}
 \label{eq:round_requirement}
\end{equation}
Indeed, $E_{\ell,R_\ell}\le\kappa_{-,t_\ell}\tau_\ell^2/4$. Evaluating~\eqref{eq:round_requirement} at $R_\ell=R$, with $D_{\ell,\rm in}=D_{\ell,\rm in}^{\rm on}$ and the calibrated error envelopes, we define
\begin{equation}
 \overline R_\ell^{\rm on}:=
 \begin{cases}
 0,&v_\ell+\widehat K_\ell\gamma_\ell^{\rm tr}
 r_{x,\ell-1}^{R_{\ell-1}}\le\tau_\ell,\\
 \displaystyle\min_{\substack{R\in\mathbb N_0\\\eqref{eq:round_requirement}\text{ holds}}}R,
 &\text{otherwise}.
 \end{cases}
 \label{eq:predictor_tolerance_test}
\end{equation}
Prediction alone suffices in the first case. Otherwise, we use the smallest count satisfying~\eqref{eq:round_requirement}, proceeding inductively from $\tau_0$. Removing conditioning on the calibration event adds at most $\zeta$ to the output failure probability.

\noindent\textbf{Step 4: Necessary uncalibrated budget.} We now compare Step 3's sufficient work with what identity prediction must do at the same tolerance. Mixing preserves the average mean, so consistent trackers and the incoming allowance give $\|\E\bar x_\ell^0-\mu_{t_{\ell-1}}\|\le e_{\ell,0}$. Step 1's surrogate errors also yield
\[
 \Delta_{\ell,0}\ge
 [\|m_{\ell-1}^\star-m_\ell^\star\|
 -\epsilon_{\ell-1}^G-\epsilon_\ell^G-\tau_{\ell-1}]_+.
\]
The latent representation gives $\|\mu_t-\alpha_tc\|\le\alpha_t r_{\rm av}$, hence $\omega_\ell\le2\alpha_{t_\ell}r_{\rm av}+\epsilon_{\Sigma,t_\ell}/a_\ell$. By the score identity in Proposition~\ref{prop:harris_geometry}, the average invariant, and~\eqref{eq:score_error}, the mean error $\mathfrak m_{\ell,k}=\E\bar x_\ell^k-\mu_{t_\ell}$ obeys
$\mathfrak m_{\ell,k+1}=q_{a,\ell}\mathfrak m_{\ell,k}+\eta_\ell\mathfrak v_{\ell,k}$,
where $q_{a,\ell}=1-a_\ell\eta_\ell$ and $\|\mathfrak v_{\ell,k}\|\le a_\ell\omega_\ell$. Unrolling and using $\|\mathfrak m_{\ell,0}\|\ge\Delta_{\ell,0}$ gives
\begin{equation}
 \|\E\bar x_\ell^{R_\ell}-\mu_{t_\ell}\|
 \ge q_{a,\ell}^{R_\ell}(\Delta_{\ell,0}+\omega_\ell)-\omega_\ell.
 \label{eq:identity_mean_lower}
\end{equation}
If every node meets the prescribed $W_2$ tolerance $\tau_\ell$, then $\|\mathfrak m_{\ell,R_\ell}\|\le\tau_\ell$. Solving~\eqref{eq:identity_mean_lower} gives $R_\ell\ge\underline R_\ell^{\rm off}$, where
\begin{equation}
 \underline R_\ell^{\rm off}=
 \begin{cases}
 \displaystyle\left\lceil
 \frac{\log((\Delta_{\ell,0}+\omega_\ell)/(\tau_\ell+\omega_\ell))}
 {-\log(1-a_\ell\eta_\ell)}\right\rceil,
 &\Delta_{\ell,0}>\tau_\ell,\\
 0,&\Delta_{\ell,0}\le\tau_\ell.
 \end{cases}
 \label{eq:full_run_lower_counts}
\end{equation}
The second case covers levels with no mean obstruction.

\noindent\textbf{Step 5: Cost comparison.} Step 3 supplies feasible calibrated counts $\overline R_\ell^{\rm on}$, whose total cost upper-bounds the minimum $C_{\rm on}(Q)$. Step 4 lower-bounds $C_{\rm off}(Q)$ through the necessary counts $\underline R_\ell^{\rm off}$. Subtracting these bounds cancels common traffic and gives
\begin{align*}
 C_{\rm off}(Q)-C_{\rm on}(Q)
 &\ge Q\sum_{\ell=1}^L c_\ell^{\rm step}
 (\underline R_\ell^{\rm off}-\overline R_\ell^{\rm on})-C^{\rm cal}\\
 &=Q\Delta_{\rm corr}-C^{\rm cal}.
\end{align*}
This proves~\eqref{eq:instance_break_even}, with strict savings when $\Delta_{\rm corr}>0$ and $Q>C^{\rm cal}/\Delta_{\rm corr}$.

\section{Experimental Details and Reproducibility}
\label{app:protocol}

\textbf{Simulated environments.} Nodes are indexed from zero. For beamforming, each site adds independent $\mathcal N(0,0.25^2)$ offsets to eight shared phases drawn uniformly from $[0.15\pi,0.65\pi]$. Snapshots add independent real and imaginary Gaussian noise, each of variance 0.05, to components of amplitude $\sqrt{0.9}$ at these phases.

For user $u=0,\ldots,7$, the offloading fraction $o$, bandwidth share $b$, and CPU share $c$ are
\begin{equation}
\label{eq:experiment_offloading_mapping}
\begin{aligned}
o&=0.5+0.45x_{u+1},\\
(b,c)&=(0.25+0.20x_{u+9},\,0.25+0.20x_{u+17}).
\end{aligned}
\end{equation}
Input sizes $z$, cycles/bit $\chi$, and rate coefficients $g$ are independent lognormal variables. Their eight log means are evenly spaced over $[-0.3,0.2]$, $[-0.1,0.4]$, and $[1.5,2.1]$, with log standard deviations $0.25,0.15,0.25$. We compute
\begin{equation}
\label{eq:experiment_offloading_cost}
\begin{aligned}
 T&=\max\!\left\{\frac{(1-o)z\chi}{1.2},\frac{oz}{bg}+\frac{oz\chi}{c(6+i)}\right\},\\
 E&=0.5(1-o)z\chi+0.15\frac{oz}{bg}.
\end{aligned}
\end{equation}
The maximum compares local and offloaded processing times.

\textbf{FLORIS environment.} We use FLORIS 4.6.4 with its Gaussian--curl hybrid configuration and 64 NREL 5-MW turbines. The array has eight columns along the wind and eight rows across it, with 630-m spacing in both directions. Wind direction, speed, and turbulence intensity are fixed at $270^\circ$, 8 m/s, and 0.06. Local labels and total-power evaluation follow~\eqref{eq:experiment_floris_objective}.

\textbf{Validation and parameter selection.} Beamforming and offloading use independent training, validation, and test channels or workloads and $\beta=4,16$, respectively. FLORIS uses independent validation and test sampling draws under the same fixed wind. Its $\beta$ is the largest value retaining at least half the local training records in effective sample size at every node:
\begin{equation}
\label{eq:experiment_floris_ess}
\frac{(\sum_j w_{ij})^2}{\sum_j w_{ij}^2}\ge\frac{N_i}{2},\qquad w_{ij}=e^{\beta(r_{ij}-\max_j r_{ij})}.
\end{equation}
Here $r_{ij}$ is a local power label and $N_i=2048$. This gives $\beta=1.61$--$1.80$ MW$^{-1}$ across seeds, limiting concentration on a few high-power records.

We test 8 or 16 geometrically spaced noise levels from 1 to 0.04. Two transition rounds span one connectivity window. Correction uses 2 or 6 mixing rounds (2 for FLORIS). Setting $\eta_\ell=c\sigma_{t_\ell}^2/n$ keeps the normalized step $n\eta_\ell/\sigma_{t_\ell}^2$ constant across levels. Validation compares $c\in\{0.04,0.12,0.3\}$. After prediction and calibration, the remaining traffic funds corrections distributed evenly across all levels or the later half (all levels for FLORIS), with remainders at the final levels. Zero corrections test prediction alone. Local uses 16 levels and 1, 4, 16, or 64 corrections per level. We rank configurations by fewest failures, highest mean worst-node quality, then lowest traffic over eight validation trials (two for FLORIS). We report coverage alongside the selected schedules' quality. We reuse a frozen schedule's test output wherever it recurs.

\textbf{Network and traffic.} Four-node rounds alternate $\{(0,1),(2,3)\}$ and $\{(1,2),(3,0)\}$, with the same edges for vectors exchanged together. Eight-node rounds analogously alternate $\{(0,1),(2,3),(4,5),(6,7)\}$ and $\{(1,2),(3,4),(5,6),(7,0)\}$. Each bidirectional float32 round costs $8nd$ bytes for calibration, $8nQd$ for tracking, and $4nQd$ for state-only mixing. Restarted consensus uses equally many state- and score-mixing rounds, each costing $4nQd$ bytes. The exact application base budget is $B_0=18432nd$ bytes.

\textbf{Statistics.} A failed run makes aggregate quality unavailable, while coverage and failure counts remain reported. For each run, the between-node variance in Table~\ref{tab:node_agreement} is
\begin{equation}
\label{eq:experiment_node_variance}
V_a=\frac1n\sum_i(a_i-\bar a)^2,\qquad \bar a=\frac1n\sum_i a_i,
\end{equation}
where $a_i$ is node $i$'s best covered objective value or coverage fraction.

\FloatBarrier

\bibliographystyle{IEEEtran}
\bibliography{references}

\end{document}